\documentclass{article} 
\usepackage{iclr2025_conference,times}

\usepackage{amsmath,amsfonts,bm}

\def\eqref#1{equation~\ref{#1}}

\def\1{\bm{1}}

\DeclareMathAlphabet{\mathsfit}{\encodingdefault}{\sfdefault}{m}{sl}
\SetMathAlphabet{\mathsfit}{bold}{\encodingdefault}{\sfdefault}{bx}{n}

\usepackage{hyperref}
\usepackage{url}

\usepackage{microtype}
\usepackage{hyperref}
\usepackage{url}
\usepackage{booktabs}
\definecolor{darkblue}{rgb}{0, 0, 0.5}
\hypersetup{colorlinks=true, citecolor=darkblue, linkcolor=darkblue, urlcolor=darkblue}
\usepackage{diagbox}
\usepackage{graphicx}
\usepackage{verbatim}
\usepackage{multirow}
\usepackage{algorithmic}
\usepackage{longtable}
\usepackage{array}
\usepackage{subfigure}
\usepackage{stfloats}
\usepackage{float}
\usepackage{balance}
\usepackage{multicol}
\usepackage{color}
\usepackage{epstopdf}
\usepackage{bm}
\usepackage{amsmath}
\usepackage{amssymb}
\usepackage{booktabs} 
\usepackage{epsfig}
\usepackage{enumitem}
\usepackage{cleveref}
\usepackage{arydshln}
\usepackage{balance}
\usepackage{xspace}
\usepackage{enumitem}
\usepackage{soul}
\usepackage{mathtools}
\usepackage{bbm}
\usepackage[ruled,linesnumbered]{algorithm2e}

\usepackage{tcolorbox}
\usepackage{caption}
\usepackage{amsmath}
\usepackage{graphicx}
\usepackage[utf8]{inputenc}

\newcommand{\vpara}[1]{\vspace{0.05in}\noindent \textbf{#1 }}

\newcommand{\plasma}{\mbox{\textsc{Plasma}}}
\newcommand{\cs}{\mbox{\textsc{C2S}}}
\newcommand{\commonsense}{\mbox{\textsc{Common2sense}}}
\newcommand{\datasetname}{\mbox{\textsc{Today}}}
\newcommand{\matres}{\mbox{\textsc{Matres}}}
\newcommand{\tracie}{\mbox{\textsc{Tracie}}}

\newcommand{\framework}{\mbox{\textsc{Bird}}}

\title{BIRD: A Trustworthy Bayesian Inference \\ Framework for Large Language Models}

\author{Yu Feng$^{1}$\thanks{ \texttt{fengyu1@seas.upenn.edu}. Code available at \url{https://github.com/CogComp/BIRD}.} \quad Ben Zhou$^{2}$\quad Weidong Lin$^{1}$\quad Dan Roth$^{1}$\\
$^1$University of Pennsylvania $^2$Arizona State University\\}

\iclrfinalcopy 
\begin{document}

\maketitle

\begin{abstract}
Predictive models often need to work with incomplete information in real-world tasks. Consequently, they must provide reliable probability or confidence estimation, especially in large-scale decision-making and planning tasks. Current large language models (LLMs) are insufficient for accurate estimations, but they can generate relevant factors that may affect the probabilities, produce coarse-grained probabilities when the information is more complete, and help determine which factors are relevant to specific downstream contexts. In this paper, we make use of these capabilities of LLMs to provide a significantly more accurate probabilistic estimation. We propose \framework{}, a novel probabilistic inference framework that aligns a Bayesian network with LLM abductions and then estimates more accurate probabilities in a deduction step. We show \framework{} provides reliable probability estimations that are 30\% better than those provided directly by LLM baselines. These estimates further contribute to better and more trustworthy decision making.
\end{abstract}
\section{Introduction}
\label{sec:intro}
\begin{figure}[t]
\centering
\label{fig:intro:examples}
\scalebox{1}{
\includegraphics[width=0.78\textwidth]{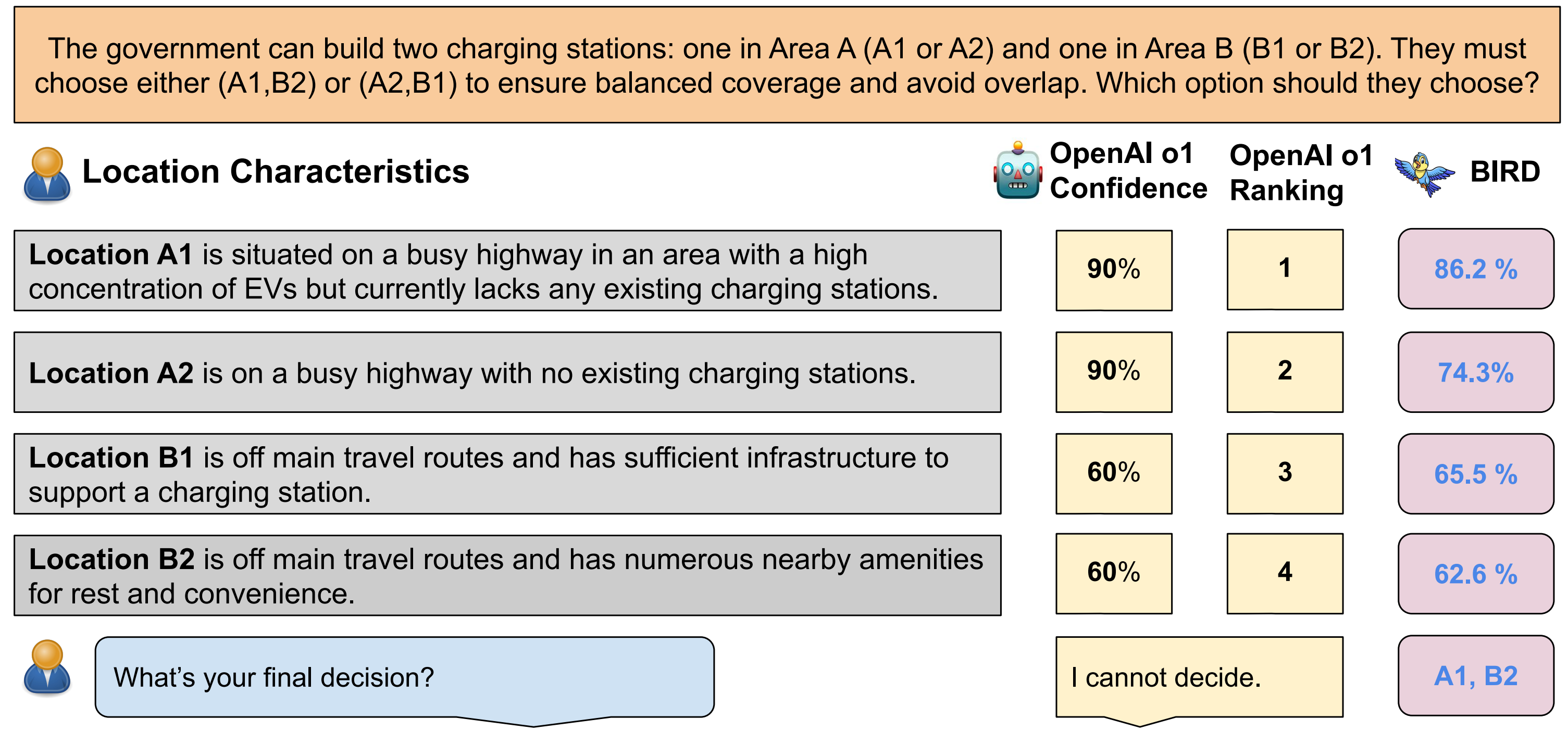}}
	\caption{\label{fig:example} \small{An example of mission-critical tasks. We first ask OpenAI o1 to separately predict the probability of building a charging station at each specific location. It estimates the same probabilities for two different conditions twice, while \framework{} (ours) successfully distinguishes the difference and can thus help the user make a more informed decision. This further demonstrates that while LLMs are capable of coarse estimations, they struggle to generate accurate probabilities. We further show through OpenAI o1 ranking that although it can successfully rank all four locations based on how likely it should be used to build a new charging station, it is still insufficient to complete the task because of ties, i.e., (1,4) and (2,3) are the same from ranking perspectives.}}
\end{figure}

Accurate, interpretable, and unbiased probabilistic estimation is crucial to artificial intelligence (AI) systems in mission-critical tasks that require trustworthy decision making. This is because, in many real-world applications, AI systems have to work with partial and incomplete observations that are insufficient for deductive and deterministic decisions~\citep{mccarthy1981some}. Consider the charging station planning scenario in Fig.~\ref{fig:example}, where we want to use large language models to decide where to build a certain number of charging stations over a region, subject to some constraints.
Since we cannot always observe all the necessary information regarding a potential location, we can only provide probability or confidence estimates on whether we should use the candidate location. If the number of planned stations is significant, we need to perform an optimization process such as ILP~\citep{roth2005integer} that maximizes the estimated probabilities of each individual station. A simple ranking of the candidates would not be sufficient in this scenario, as it may yield a suboptimal solution illustrated in Fig.~\ref{fig:example}. However, current LLMs are not ideal for such probability estimation tasks since 1) LLMs tend to be inaccurate and overconfident when estimating numerical confidences and probabilities~\citep{xiong2024can} and 2) they cannot provide an out-of-the-box interpretable and controllable process on how their estimates are derived~\citep{li-etal-2024-deceptive}. This prevents us from using LLMs' decision-making capabilities in large-scale automatic decision-making tasks. Consequently, there is a need to find ways to infer better probabilities from these models.

In this work, we propose \framework{} (\textbf{B}ayesian \textbf{I}nference f\textbf{r}om Abduction and \textbf{D}eduction). 
\framework{} builds on three observations. The first observation is that although LLMs cannot produce convincing numerical probabilities directly, they can generate neutral and comprehensive real-world factors relevant to these probabilistic estimates. Moreover, we observe that they can make relatively correct coarse probability estimations when using these factors as contexts that provide more complete information. 
The last observation is that LLMs can decide how specific contexts and conditions are related to these factors via entailment classifications.
Together, these observations motivate \framework{}, where we approximate a Bayesian network~\citep{friedman1997bayesian} using LLM-generated factors, optimize its internal conditional probabilities by sampling LLM coarse predictions on combinations of values of all the factors, and infer more accurate probabilities for specific downstream conditions by leveraging LLMs to map conditions to factors.

\begin{figure*}[ht]
\label{fig:modeling:overview}
\centering
\scalebox{0.9}{
	
	\includegraphics[width=0.98\textwidth]{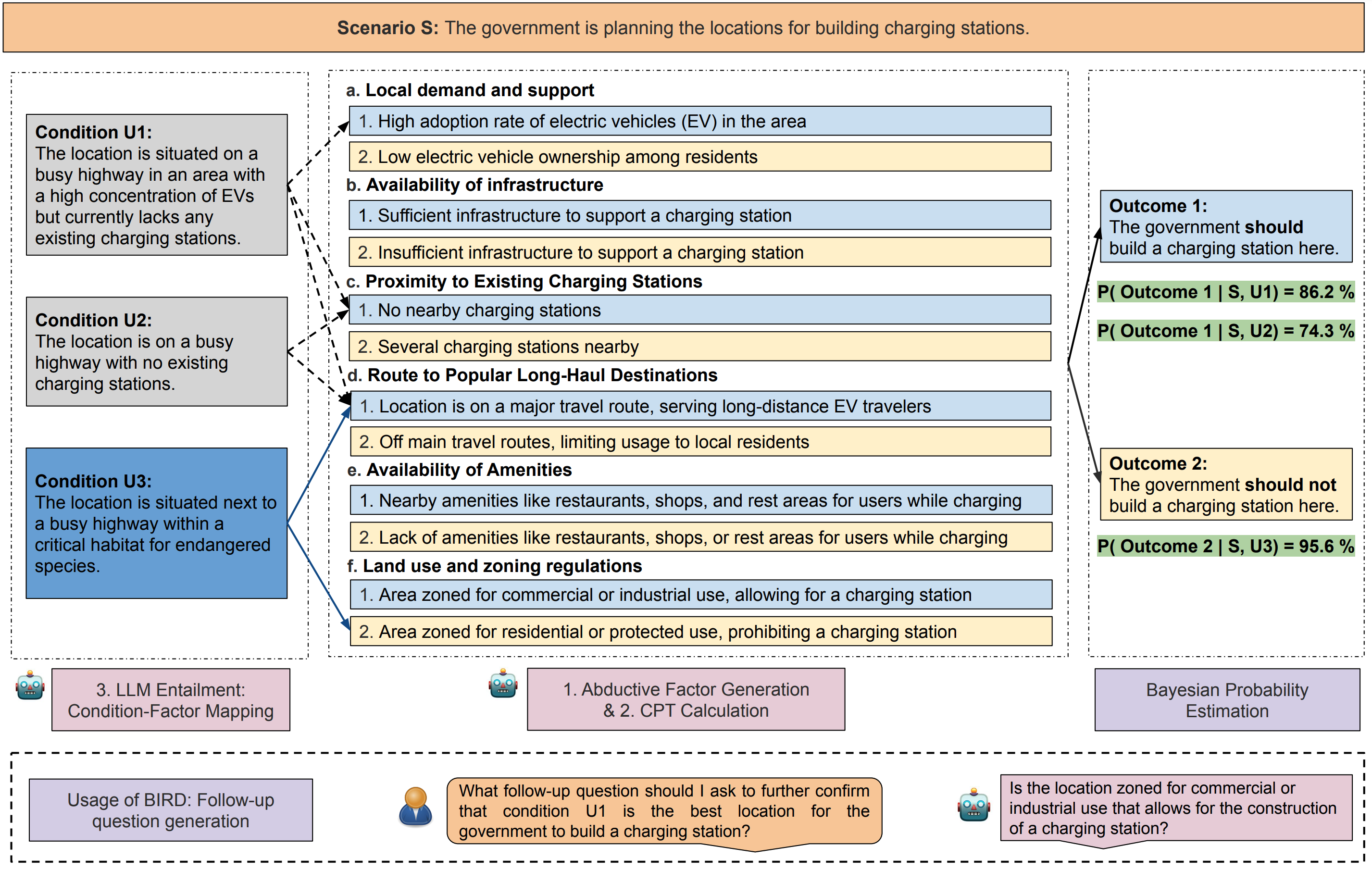}}
 	\caption{\label{fig:main}\small Overview of \framework{}. Given a scenario $S$, LLMs generate the factors $F$ (a/b/...) with potential values ($f_1 \in$ \{a.1,a.2\}). \framework{} approximates a Bayesian network parameterized by $\mathbb{P}(O_i|f_j)$, and optimizes by sampling LLM coarse predictions on $\mathbb{P}_\mathrm{LLM}(O_i|f), f \in \mathcal{F}$ ($\mathcal{F}$ is the set of all value combinations of $F$, e.g., $f={\rm ( a.1,b.2,c.1,d.2,e.1,f.2)}$), and minimizing the distributional distance between an approximated $\mathbb{P}_\mathrm{estimated}(O_i|f)$ and $\mathbb{P}_\mathrm{LLM}(O_i|f)$. At inference time, each context $C$ (S+U1/S+U2/...) is mapped to some $f_j$ via entailment, and a probability is derived using $\mathbb{P}_\mathrm{estimated}$. \framework{} can further generate follow-up questions.
  }
\end{figure*}

As illustrated in Fig.\ref{fig:main}, we formulate the probability estimation problem as deciding $\mathbb{P}(O_i|C)$, where $O_i$ is a potential outcome (e.g., the government should/should not build 
 a charging station here) and $C$ is the downstream condition or context (e.g., the location is on a busy highway) under a decision-making scenario $S$ (e.g., deciding whether to build a charging station at the location). 
Given $S$, \framework{} first generates a comprehensive set of real-world factors and values $f_1, f_2, ..., f_n$ that may contribute to the probabilistic estimations. Then, it approximates a Bayesian network structure using the generated factors and the outcomes $\{O\}$. 
We propose a novel method for estimating the conditional probability table (CPT) for the outcome, i.e., $\mathbb{P}(O_i|f),f \in \mathcal{F}$ where $\mathcal{F}$ is the product space of value sets for all factors.\footnote{$\mathcal{F}$ is the set of all possible combinations of different assignments of values to all factors (e.g., selecting a value from each of the factors a/b/c/d/e/f in Fig.~\ref{fig:main}). Refer to \S\ref{turn1}.} Specifically, we assume a set of learnable parameters $\mathbb{P}(O_i|f_j)$ representing the conditional probability of each possible outcome given a single variable $f_j$.
We then sample $f \in \mathcal{F}$ and acquire LLM's coarse probability estimations $\mathbb{P}_{LLM}(O_i|f)$. 
Next, we approximate the conditional probability $\mathbb{P}_{estimated}(O_i|f)$ in the CPT using the learnable parameters $\mathbb{P}(O_i|f_j)$ based on a derived approximation formula and optimize these parameters by minimizing the distributional distance between $\mathbb{P}_{estimated}(O_i|f)$ and the LLM's coarse predicted probability $\mathbb{P}_{LLM}(O_i|f)$. From a high level, this abduction step aligns language models' internal probability estimation with Bayesian constraints and has better error tolerance. The resulting Bayesian network is generic to all possible conditions in the scenario $S$.
In the following deduction step, we use the trained Bayesian parameters to estimate the probability of any possible conditions and observations $\mathbb{P}(O_i|C)$. To do this, we employ an entailment process to determine $\mathbb{P}(f_i|C)$, as shown by ``LLM Entailment'' in Fig.~\ref{fig:main}, and the outcome we are interested in, $\mathbb{P}(O_i|C)=\sum_{f\in \mathcal{F}} \mathbb{P}(O_i|f)\mathbb{P}(f|C)$.

As we show in \S\ref{sec:experiment}, the inferred probabilities from \framework{} are more reliable, consistent with human judgments on 30\% more instances than those directly predicted by LLMs. We further show that \framework{} probability predictions are good enough to be directly used for decision making (i.e., by computing an argmax on the predicted probabilities of each label for inference), outperforming standard chain-of-thought (CoT)~\citep{wei2022chain} on several reasoning benchmarks. More importantly, \framework{} achieves better probability estimations by optimizing an interpretable Bayesian network with natural-language-based variables. There are two additional benefits of \framework{}. First, we show that \framework{}-aligned probabilities can serve as more accurate and fine-grained supervision signals for training smaller models. In addition, we demonstrate that \framework{}'s factor generation process is neutral and unbiased to the specific conditions, so it can be used to generate better follow-up questions to improve decision-making confidence. Combining these findings, we show that \framework{} improves trustworthiness --- a crucial ability for mission-critical applications such as medical chatbots, where we need reliable decisions supported with enough evidence for high-confidence predictions.

\section{Related Work}
\vpara{Direct Inference in LLMs.} Our paper is first related to LLM direct inference methods~\citep{wei2022chain, wang2023selfconsistency}, and decomposition-based reasoning~\citep{ wolfson-etal-2020-break, tafjord-etal-2022-entailer, zhou-etal-2022-learning-decompose, kassner-etal-2023-language}. Compared to these works, \framework{} considers using abduction and deduction to facilitate probabilistic induction, which aligns with the slow-thinking systems~\citep{McGlynn2014ThinkingFA, Lin2023SwiftSageAG}. Our work is also related to chain-of-thought via latent-variable inference. Given a question-answer pair $(X,Y)$, existing methods ~\citep{hoffman2023training,hu2024amortizing} aim to find latent chains of thought – token sequences $Z$ that contribute the most to the conditional likelihood. While \framework{} has a more structured $Z$ and focuses more on the proper probability estimation instead of merely finding the optimal $Z$.

\vpara{Uncertainty Estimations and Decision making under Uncertainty in LLMs.} 
Most current works focus more on sampling to access the model’s uncertainty of factual knowledge~\citep{kuhn2023semantic,yadkori2024believebelievellm,lin2024generatingconfidenceuncertaintyquantification}. Some works verbalize the uncertainty of LLMs~\citep{tian-etal-2023-just,xiong2024can}. However, \citet{xiong2024can} shows that LLMs are often overconfident when directly verbalizing their confidence. \cite{pmlr-v235-hou24b} is most related to us, as it also incorporates the intuition of decomposition, but it focuses more on addressing aleatoric uncertainty due to input ambiguity without rigorous probability calculation. This work is also related to decision making with uncertainty~\citep{ren2023robots, han2024uncertaintyaware, liu2025dellma} while all methods use LLM's direct uncertainty estimation such as token logits or verbalization. \framework{} introduces external reliable Bayesian modeling to more accurately calibrate uncertainty in LLMs for decision making. 

\vpara{Probabilistic Inference with LMs.} 
\cite{chen-etal-2020-uncertain} is one of the first papers to estimate a probability out of an NLI problem while with no explanation. ~\cite{ozturkler-etal-2023-thinksum} follows a similar intuition to ours, where the model first thinks by retrieving relevant associations, and then conducts probabilistic reasoning. \cite{lew2020leveraging,wong2023wordmodelsworldmodels} discuss translating from natural language to the probabilistic language of thought. However, none of the papers tackle the real-world complex settings under incomplete information as comprehensively as we do with the level of interpretability to users of any level.

\section{Modeling}
\label{sec:model}

\label{sec:preview}
\subsection{Bayesian Inference from Abduction and Deduction}
\label{turn1} 

On a high level, given a decision $Y$ and partial information $X$, instead of direct inference using LLMs $({\rm induction}: X \rightarrow Y)$, 
\framework{} conceptualizes the general scenario behind $X$ into intermediate factors $({\rm abduction}: X \rightarrow Z)$, and then fits a Bayesian model with $Z$. The model then estimates probabilities for $Y$ $({\rm deduction}: X, Z \rightarrow Y)$. 
Our problem setting is a generic contextual inference problem of the type inspired by \citet{feng-etal-2023-generic}.
Given the context $C$, which includes a general scenario $S$ and a corresponding downstream condition $U$, our task is to estimate $\mathbb{P}(O_i|C), i=1,2, C = (S, U)$, where $\{O\}$ is denoted as $O_1$ and $O_2$. We decompose the overall context $C$ into the general scenario $S$ and downstream conditions $U$ because we want to build a neutral intermediate space $Z$ based on only $S$, so that the probability estimation is unbiased towards any specific conditions. This would mitigate models relying on incorrect signals, such as \textit{charging stations are normally built in higher-income neighborhoods}, but more generic factors like \textit{accessibility} for trustworthy decision making.

Instead of directly making a prediction, we approximate a controllable and interpretable inference process based on explicit Bayesian modeling~\citep{reason:GelCarSteRub95} containing natural-language-based variables. 
For a scenario $S$, we conceptualize $N$ factors $\{F_j\}_{j=1}^N$ based on $S$ that are relevant to the outcome $O$, which can take on two possible values, $O_i, i=1,2$, similar to the structure found in Bayesian networks~\citep {10.5555/1795555}. Here the real-world factors and the outcome are equivalent to the variables in the network. Therefore, the choice between outcomes $O_1$ and $O_2$ depends on the factors, with each discrete factor $F_j$ having a set of possible values as $\mathcal{F}_j$, containing all possible details of the factor required for deciding between $O_1$ and $O_2$. 
We use $\mathcal{F}= \prod_{j=1}^N \mathcal{F}_j$ to denote the product space of value sets for all factors. For an instance in the space, each factor $F_j$ should be attributed a specific value $f_j$. We assume such derived factors are complete at this stage. That is, they include all the possible individual factors/angles that may affect the distribution of the outcomes.\footnote{Note that our framework can later accommodate any additional factors by selectively recalculating specific parts as needed.} As a result, $\mathcal{F}$ is a complete information space for decisions. As in Fig.~\ref{fig:main}, (factor a value 1 $a_1$, factor b value 1 $b_1$, factor c value 1 $c_1$, factor d value 1 $d_1$, factor e value 1 $e_1$, factor f value 1 $f_1$) is an element $f$ in the product space $\mathcal{F}$ where there are $2^6$ elements in total. We then calculate the conditional probability table (CPT) as $\mathbb{P}(O_i|f),\forall f \in \mathcal{F}, f = (f_1,f_2,...,f_N)$.

We assume 1) $\{F_j\}_{j=1}^N$ is complete, so $O$ does not depend on any other intermediate factors, and 2) only one value can be assigned for each factor for a given $C$.
The predictive probability for the outcome is thus obtained by marginalizing over the product space $\mathcal{F}$ for a context $C$ (details in \S\ref{appendix:derivation}): 
\begin{equation}
\label{eq:main}
\mathbb{P}(O_i|C) = \sum_{f\in \mathcal{F}} \mathbb{P}(O_i|f)\mathbb{P}(f|C)
\end{equation}
where $i = 1,2, C = (S, U)$, $ \forall f \in \mathcal{F}, f = (f_1,f_2,...,f_N), f_j \in \mathcal{F}_j, j=1,...,N$. $\mathcal{F}$ is the product space of value sets for all factors. $f$ is a possible instance of complete information in the space. 

\vpara{Assumption on Conditional Independence of Variables.} To provide relaxation for real-world applications, and since the factors should cover as many different aspects as possible, we assume factors are conditionally independent given the context. Therefore, 
\begin{equation}
\label{eq:independent}
\mathbb{P}(O_i|C) = \sum_{f\in \mathcal{F}} \mathbb{P}(O_i|f)\prod_{j=1}^N\mathbb{P}(f_j|C)
\end{equation}

In the following sections, we describe the missing implementation details of the framework: (1) the generation of the factors $\{F_j\}_{j=1}^N$ (\S\ref{turn2}); (2) the assignment of the conditional probability table $\mathbb{P}(O_i|f)$ (\S\ref{turn3}), and (3) the computation of observations $\mathbb{P}(f_j|C), j=1,...,N$ (\S\ref{turn4}).

\subsection {Abductive Factor Generation} \label{turn2} 
Large language models, although they cannot solve some complicated cases, are shown to possess parametric knowledge that enables them to solve more common cases that can be directly found in their memory~\citep{kambhampati2024position,li-etal-2024-deceptive, zhou2024conceptual}. Our method for generating the factors follows this similar intuition.
For a scenario $S$, we first do abstractions with LLMs to derive $N$ 
factors $\{F_j\}_{j=1}^N$. The most straightforward approach involves utilizing LLMs to directly obtain a list of factors and their potential values. However, direct generation is effective but not sufficiently comprehensive, and we want to fully utilize past similar occurrences that the model memorizes in their parametric knowledge. 
 
Therefore, to encourage models to fully explore the range of alternatives, we divide the task into two stages: 1) generating sentences describing familiar situations that would increase the likelihood of an outcome, and 2) summarizing these sentences into factors with corresponding values. We show the prompts in Appendix Fig.~\ref{fig:factor_added_info} and Fig.~\ref{fig:factor_structure}. 
We further utilize LLMs to perform classification to assess how the value of each generated factor affect the decision, i.e., deciding the value supports which outcome, so that we can filter out factors that are unrelated to the outcomes. 
The prompt is in Appendix Fig.~\ref{fig:factor_outcome}. 
In the example depicted in Fig.~\ref{fig:main}, a factor like ``The color of the vehicle'' will be eliminated since all potential values of this factor consistently support the same outcome, i.e., neutral.

\subsection {Assignment of $\mathbb{P}(O_i|f)$ as a constrained optimization problem} \label{turn3} 

We now calculate each $\mathbb{P}(O_i|f), i=1,2, \forall f \in \mathcal{F}$ in the conditional probability table of the Bayesian network. Since we do not have any observed data for approximation and LLMs alone cannot directly give an accurate assignment on $\mathbb{P}(O_i|f)$, we adopt the formula proposed in \citet{bordley1982multiplicative}, which derived a version of the logarithmic opinion pool using axioms from the theory of additive conjoint measurement. This formula is proved to be the only continuous (over the open interval (0, 1)) group probability formula under the assumption detailed in \S\ref{appendix:assumption} that the decision maker has an intuitive weak ordering on the set of factor odds ratios and a ``noninteraction'' property of factors. Our setting fits the assumption as the weak ordering of decisions is the most fundamental condition of consistency in decision making and we assume the conditional independence of factors. Therefore,
\begin{equation}
\mathbb{P}(O_i|f) =\frac{\prod_{j=1}^N(\frac{\mathbb{P}(O_i|f_j)}{\mathbb{P}(O_i)})^{w_j}\mathbb{P}(O_i)}{\prod_{j=1}^N(\frac{\mathbb{P}(O_i|f_j)}{\mathbb{P}(O_i)})^{w_j}\mathbb{P}(O_i) + \prod_{j=1}^N(\frac{1-\mathbb{P}(O_i|f_j)}{1-\mathbb{P}(O_i)})^{w_j}(1-\mathbb{P}(O_i))} 
\end{equation}
To determine the weights $w_j$,\footnote{Weights here refer to a set of ``hyperparameters'' that decides how highly the final decision regards each factor and how correlated the factors are considered, not the learnable parameters in our Bayesian network.} 
, we adopt the second case scenario mentioned in \citet{bordley1982multiplicative}, i.e., $\sum_{j=1}^N w_j > 1$. In this case, the decision maker with complete information is more certain than the decision maker with only one piece of information through acquiring all different pieces of information from different factors. Since there isn't a priori information that suggests another choice exists, we use equal weight $w_j=1, j = 1,..., N$ and assume $\mathbb{P}(O_i) = 50\%, i = 1,2$. The final estimation formula can be rewritten as:  
\begin{equation}
\label{eq:estimation}
\mathbb{P}_{\rm estimated}(O_i|f) = \mathbb{P}_{\rm estimated}(O_i|f_1,f_2,...f_{N})  \\
\approx \frac{\prod_{j=1}^N\mathbb{P}(O_i|f_j)}{\prod_{j=1}^N\mathbb{P}(O_i|f_j) + \prod_{j=1}^N(1-\mathbb{P}(O_i|f_j))} \\
\end{equation}

\vpara{Constrained optimization for estimating $\mathbb{P}(O_i|f)$.}\label{learning_algorithm} 
From Eq.~\ref{eq:estimation}, it is not hard to see that each $\mathbb{P}(O_i|f), \forall f \in \mathcal{F}$ can now be approximated using only $\mathbb{P}(O_i|f_j), \forall f_j \in \mathcal{F}_j, j = 1, ..., N$, which we view as learnable parameters. 
As discussed in \S\ref{sec:intro}, LLMs can perform relatively correct coarse estimation under complete information, which refers to contexts with all factors' values present (i.e. $f \in \mathcal{F})$. Consequently, we prompt the LLMs to output verbalized probability estimations for outcome $O_i$ given $f$.
This approximates LLM's predictions of $\mathbb{P}_{LLM}(O_i|f)$.
With $\mathbb{P}_{\rm estimated}$ and $\mathbb{P}_{LLM}$, we can now form a constraint optimization problem that finds the values of parameters $\mathbb{P}(O_i|f_j)$
that minimize the distributional distance between the Bayesian estimated probabilities $\mathbb{P}_{\rm estimated}(O_i|f)$ and LLM estimated probabilities $\mathbb{P}_{LLM}(O_i|f),\forall f \in \mathcal{F}$, while conforming to the structure in Eq.~\ref{eq:estimation} and preliminary assessments about each $\mathbb{P}(O_i|f_j), \forall f_j \in \mathcal{F}_j, j = 1, ..., N$ provided by the LLM classification as mentioned in \S\ref{turn2}. 

\vpara{Learning algorithm for constrained optimization to estimate $\mathbb{P}(O_i|f)$.} 
\label{learning_algo} We propose a gradient-descent algorithm for this optimization problem.  
The detailed algorithm can be found in \S\ref{appendix:algorithm}, and we give a general description here.
As the parameter initialization,  we utilize the LLM classification as mentioned in \S\ref{turn2}, where we direct the LLMs to ascertain which outcome is more likely supported by the value of each factor:
\begin{equation}
\label{eq:initial_prob}
\mathbb{P}_{\rm init}(O_i|f_j)  = \left\{ 
    \begin{array}{lc}
       75\% & f_j \  {\rm supports}  \ {\rm outcome} \ i\\
       50\% & f_j \ {\rm is} \ {\rm neutral} \\
       25\% & f_j \  {\rm supports}  \ {\rm opposite} \ {\rm outcome} \ \neg i \\
    \end{array}
\right.
\end{equation}
For the training data,  
given an instance of complete information $f^{'}= (f_1^{'},f_2^{'},...,f_N^{'})$, 
we prompt LLM to output a verbalized probability score for each outcome $\in$ \{very unlikely, unlikely, somewhat unlikely, neutral, somewhat likely, likely, very likely\} which we use as $\mathbb{P}_{LLM}(O_i|f^{'})$. The detailed prompt can be found in Appendix Fig.~\ref{fig:llm_training_signal}. We then assign a specific probability value for each coarse estimation, i.e., very unlikely: 0\%, unlikely: 20\%, somewhat unlikely: 40\%, neutral: 50\%, somewhat likely: 60\%, likely: 80\%, very likely: 100\%, as supervision signals to learn the correct value of $\mathbb{P}(O_i|f_j^{'}), j = 1,..., N$ for each $f_j^{'}$ in $f^{'}$ using Eq.\ref{eq:estimation}. We randomly sample 128 instances in the space of complete information as the training data for each scenario. Note that we manually assign probability values for LLM coarse decisions, but they can be changed as hyperparameters.

We adopt two loss functions: 1) we use the regular mean squared error loss for optimization and denote the loss as $\ell_{\rm MSE}(\mathbb{P}_{\rm estimated}(O_i|f^{'}), \mathbb{P}_{\rm LLM}(O_i|f^{'}))$; 2) we use the margin ranking loss to ensure the final trained individual probability preserve the original direction of its supported outcome and denote the loss as $\ell_{MR}$: 
\begin{equation}
\begin{split}
\ell_{\rm MR} &= {\rm max}(0, -y_{\rm target}(O_i|f_j) (\mathbb{P}_{\rm trained}(O_i|f_j)- 0.5) + \epsilon) \\
y_{\rm target}(O_i|f_j) &= \mathrm{sgn}(\mathbb{P}_{\rm init}(O_i|f_j) - 50\%)\\
\mathbb{P}_{\rm trained}(O_i|f_j) &= \sum_{f^{*} \in \mathcal{F}^*} \mathbb{P}(O_i|f^{*} \cup f_j ) \mathbb{P}(f^{*}  | f_j ) = \frac{1}{{\rm card}(\mathcal{F}^*)}\sum_{f^{*} \in \mathcal{F}^*}\mathbb{P}(O_i|f^{*} \cup f_j )
\end{split}
\end{equation}
$\forall f_j \in \mathcal{F}_j, j = 1, ..., N$. We apply probability marginalization to get the trained individual probability and approximate it with the mean values since there are no explicit priors. We use $\mathcal{F^*}$ to denote the product space of the remaining factors except $\mathcal{F}_j$, which is a subspace of $\mathcal{F}$ and $f^{*} = (f_1, f_2, ..., f_{j-1}, f_{j+1}, ..., f_N), f^{*} \in \mathcal{F^*} $. Therefore, the final loss function we use for training is defined as follows:
\begin{equation}
\label{eq:loss}
\mathcal{L} = \ell_{\rm MSE} + \alpha \ell_{\rm MR}
\end{equation}
where $\alpha$ balances the two losses. We use SGD as the optimization function. As a result, we can estimate any $\mathbb{P}(O_i|f), \forall f \in \mathcal{F}$ with the learned $\mathbb{P}(O_i|f_j), \forall f_j \in \mathcal{F}_j$ with Eq.\ref{eq:estimation}.

\vpara{Customization to human preference.} A benefit of our design is that we can explicitly ask for human preferences on how a particular factor value will affect the outcome, i.e., manually setting $\mathbb{P}(O_i|f_j)$. Such preferences are desirable in certain applications with known constraints, and they can be easily integrated in Eq.\ref{eq:estimation} by replacing $\mathbb{P}(O_i|f_j)$ with the human preferred value.

\subsection {Computation of $\mathbb{P}(f_j|C)$ with LLM Entailment} \label{turn4} 

This section discusses how we compute $\mathbb{P}(f_j|C)$. As noted in \S\ref{turn1}, $C$ is the context containing the general scenario $S$, and a specific additional condition $U$. Given $C$, we employ an entailment task formulation to find which factors and their corresponding values are implied or mapped by $C$.
Specifically, we adopt the prompt in Appendix Fig.~\ref{fig:entailment} where we directly ask if the context entails a value from a factor. We conduct another round of self-correction using the prompt in Appendix Fig.~\ref{fig:entailment_reeval} to check if the model believes all the previously identified values are actually entailed.

We define the notion of observed factors, which refers to a subset of factors among all factors that are being implied by the context. For example, as shown in Fig.~\ref{fig:main}, context $S+U_1$ implies $a_1$, $c_1$, and $d_1$, $S+U_2$ implies $c_1$ and $d_1$, while the additional condition $S+U_3$ implies $d_1$ and $f_2$. We assume only one value can be chosen per factor. We write the product value space of the implied factors as $\mathcal{F'}=\prod_{m=1}^M{\mathcal{F}_{j_m}}$, which is a subspace of $\mathcal{F}$. We define the conditional probabilities of the values of these observed factors to be 1 if the value is implied by $C$, a 0 if not:

\begin{equation}
\begin{split}
\mathbb{P}(f_{j_m}|C) = \left\{ 
    \begin{array}{lc}
       1 & f_{j_m} = f_{j_m}^\ast, f_{j_m} \in \mathcal{F}_{j_m}  \\
       0 & f_{j_m} \neq f_{j_m}^\ast, f_{j_m} \in \mathcal{F}_{j_m} \\
    \end{array}
\right.
\end{split}
\end{equation}
Where the implied $f_{j_m}^\ast \in \mathcal{F}_{j_m}, \forall m=1,...,M$.
For the unobserved factors, we have no information regarding which value to be selected. In order to be neutral and unbiased, we assume that, in an unobserved factor, each value has an equal probability of being selected.
We use ${F'}^{\perp}= \prod_{m=M+1}^{N}\mathcal{F}_{j_m}$ to denote the space of unobserved factors, which is the orthogonal complement space of $\mathcal{F'}$ in  $\mathcal{F}$, 
\begin{equation}
\label{eq:final}
\mathbb{P}(f_{j_m}|C) = \frac{1}{{\rm card}(\mathcal{F}_{j_m})} ,  f_{j_m} \in \mathcal{F}_{j_m}
\end{equation}
Where $\forall m=M+1,...,N$. We can therefore calculate the final probability as presented in Eq.\ref{eq:main}, utilizing the formulations provided in Eqs.\ref{eq:independent} through \ref{eq:final}.

\section{Experiments}
\label{sec:experiment}
\subsection{Datasets and Settings}
\label{sec:datasetsandsettings}
We convert three datasets focusing on reasoning and planning to our problem setting including $O$, $S$, and $U$, as described in \S\ref{turn1}. We detail this process in \S\ref{appendix:dataset}.
\textbf{\commonsense{}}~\citep{singh-etal-2021-com2sense} focuses on commonsense reasoning, and we gather 216 scenarios and 3822 instances, where 9 additional conditions on average support each outcome. \textbf{\plasma{}}~\citep{brahman2023plasma} focuses on planning, and we gather 279 scenarios and 1395 instances, where 5 additional conditions support the less common outcome. \textbf{\datasetname{}}~\citep{feng-etal-2023-generic} focuses on temporal reasoning, where there are 1000 instances. Among the three datasets, the most challenging task is \datasetname{} and the easiest is \commonsense{}, judged by baseline model performances. The complexity of the scenario and conditions is highest in \datasetname{}. The conditions support the less common outcome in \plasma{} and \commonsense{} presents general decision making with common sense.

We conduct our experiments with Llama-2-70b-instruct and Llama-3.1-70b-Instruct.\footnote{\url{https://huggingface.co/upstage/Llama-2-70b-instruct} and \url{https://huggingface.co/meta-llama/Llama-3.1-70B-Instruct}} 
For the factor generation task in \S\ref{turn2}, we generate 10 sentences for each outcome before summarization. 
For the LLM classification task for factor-outcome mapping, we generate 3 times and conduct a majority vote.
For the LLM entailment task in \S\ref{turn4}, we sample 3 times and conduct a majority vote with 1 round of self-reflection. 
We set learning rate = 1e-02, epoch = 20, batch size = 4, $\epsilon$ = 0, $\alpha$ = 10 for the learning algorithm mentioned in \S\ref{turn3}.

\subsection{Intrinsic Evaluations of the \framework{} Framework}
\label{sec:intrinsic-eval}

\begin{table*}[t]
\centering
\small
{
\begin{tabular}{lcccc}
\toprule
Model & Different$_1$ & Different$_2$ & Same & Average\\
\cmidrule(lr){1-1}\cmidrule(lr){2-2}\cmidrule(lr){3-3}\cmidrule(lr){4-4}\cmidrule(lr){5-5}
Random Guessing & 0.333 & 0.333&  0.333&  0.333 \\
\cmidrule(lr){1-5}
GPT3.5 CoT  & 0.306 &  0.306 &  0.242 & 0.283 \\
GPT4 CoT & 0.312& 0.357&0.216 & 0.289 \\
Llama2-70b Instruct Logit & 
0.263 & 0.228 &0.205 & 0.228 \\
Llama2-70b Instruct Vanilla	& 0.375 & 0.333 &  0.243 &  0.311\\
Llama2-70b Instruct  CoT & 0.315 &  0.323 &  0.254 & 0.294\\
Llama3.1-70b Instruct Logit & 0.300 &  0.282 & 0.242 & 0.269
 \\
Llama3.1-70b Instruct Vanilla & 0.365 &  0.301 &  0.251 & 0.303\\
Llama3.1-70b Instruct CoT	& 0.373 & 0.351 & 0.187 & 0.309\\

\cmidrule(lr){1-5}
Llama2-70b Instruct EC* & 0.530 & 0.529 &  0.207 & 0.503\\
Llama3.1-70b Instruct EC* &0.535 & 0.538 & 0.286 & 0.511\\

GPT4 EC* & 0.588 & 0.533 &  0.300 & 0.540\\ 
\cmidrule(lr){1-5} 
Llama3.1 \framework{} (\textit{ablation w 1/2 assumption}) & 0.527& 0.532 &0.196 & 0.480\\
Llama3.1 \framework{} (\textit{ablation w 1/n assumption}) & 0.572 & 0.584 & 0.272 & 0.532\\
Llama3.1 \framework{} (\textit{ablation w fixed initial prob}) & 0.614 & 0.597 & 0.337 & 0.568\\
\cmidrule(lr){1-5} 
Llama2 \framework{} (\textit{ours w optimized prob}) & \textbf{0.614} & 0.624 & \textbf{0.450} & \textbf{0.592}\\
Llama3.1 \framework{} (\textit{ours w optimized prob}) &0.612& \textbf{0.625} &0.382 & 0.588 \\

\bottomrule
\end{tabular}
}
\caption{Preference-based pairwise evaluation using F1 score for predicted probabilities across various model settings. Different$_1$ indicates the F1 score for a category where $C_1$ supports the gold outcome better than $C_2$, and vice versa for Different$_2$. Same refers to the F1 score of the category where $C_1$ and $C_2$ equally support the outcome. Average refers to calculating F1 metrics globally by counting the total true positives, false negatives, and false positives. EC* settings see both $C_1$ and $C_2$ at the same time instead of predicting individual probabilities, hence not directly comparable.} 
\label{tab:main}
\end{table*}

\vpara{Reliability of \framework{}'s Estimated Probabilities.}
\label{exp:prob} 
We first propose an evaluation method that directly benchmarks the quality of probability estimations. We adopt a similar setting as mentioned in \S\ref{turn1}. Given a scenario $S$ and two potential outcomes $O_1$ and $O_2$, we find two specific additional conditions $U_1$ and $U_2$, both supporting $O_1$ over $O_2$.\footnote{Here $O_1$ will be the gold outcome. During evaluation, the two conditions can also support $O_2$ over $O_1$.} $S+U_1$ and $S+U2$ will form two contexts $C_1$ and $C_2$, respectively. The goal of this task is to evaluate if models can assign fine-grained probability estimations, that is, sufficient to distinguish which context between $C_1$ and $C_2$ \textit{\textbf{better}} supports $O_1$ over $O_2$ or if both $C_1$ and $C_2$ \textit{\textbf{equally}} support $O_1$ over $O_2$, by predicting probability values of $P(O_i|C1)$ for $C_1$ and $P(O_i|C2)$ for $C_2$ \textbf{\textit{individually}}.\footnote{For proper evaluation, systems predict probabilities one at a time, without seeing other contexts.} That is, assuming $C_1$ better support $O_1$ over $O_2$, we expect $P(O_1|C_1)>P(O_1|C_2)>P(O_2|C_2)>P(O_2|C_1)$, and vice versa. If $C_1$ and $C_2$ equally support $O_1$ over $O_2$, we should see $P(O_1|C_1) =P(O_1|C_2)>P(O_2|C_1)=P(O_2|C_2)$. This task is significantly more difficult than providing two contexts that support different outcomes, as systems will have to understand the nuances in probability estimation. 

We collect human gold labels for which context better supports the gold outcome with rigorous quality control using MTurk. We explain our choice of this relative evaluation over asking human annotators to provide absolute probabilities from a psychological perspective, along with the details of human annotation in \S\ref{appendix:humaneval}. We evaluate on 350 randomly sampled instances from \commonsense{} when the conditions can be mapped to at least one factor and therefore \framework{} does not produce ``unknown''\footnote{Although the setting  may seem to favor our proposed method, it is a fair comparison, because ``unknown'' predictions are label-agnostic, and we can use the same direct inference approach, such as CoT, for such cases.}.
We report the F1 score as the evaluation metric. We further explain how the F1 score is computed in \S\ref{appendix:humaneval}. 

We compare the following methods that directly acquire probability estimations from LLMs as baselines. 1) \textbf{Vanilla}~\citep{wang2023selfconsistency}: models directly verbalize the estimated probabilities. 2) \textbf{Logit}: we select the normalized token probability of the decision. 3) \textbf{CoT}~\citep{wei2022chain}: models output a reasoning process followed by verbalized estimated probabilities. 4) \textbf{EC}: models select which additional condition better supports the outcome without outputting any probabilities. Note that this formulation is unfair to \framework{} and other baselines that individually estimate the probabilities. To demonstrate the effectiveness of our learning algorithm in \S\ref{turn3}, we propose two ablation baselines that use other methods to find the conditional probabilities in the Bayesian network: 1) \textbf{1/2 and 1/N Assumptions}: We adopt two neutral assumptions without LLM classifications for CPT calculation. 1/2: $\mathbb{P}(O_i|f)$ is 1 only if all factors' values, except for any neutral ones, support $O_i$; if not, 0. 1/N: $\mathbb{P}(O_i|f)$ is calculated as the percentage of factors whose values support outcome
$O_i$. 2) \textbf{Fixed Initial Prob}: We utilize the initial probability of Eq.\ref{eq:initial_prob} to directly estimate $\mathbb{P}(O_i|f)$ with Eq.\ref{eq:estimation}. The details of the baselines are presented also in \S\ref{appendix:baseline}. 

Table~\ref{tab:main} lists the results. Our framework exhibits superior alignment with human preference, with both models achieving an F1 of 59\%, more than 30\% higher than GPT-4. Such results demonstrate that \framework{} produces reliable probability estimation without additional human supervision. Notably, such probability estimation is done by judging each condition separately (i.e., \framework{} assigns one probability to condition 1, and later another probability to condition 2). We find that vanilla language models do much better when both conditions are provided at the same time (i.e., the EC setting) where the model can perform argmax operations. However, even under this setting that is unfair to our framework, \framework{} still outperforms GPT-4 by 5\%. 
All results suggest that vanilla LLMs cannot properly estimate probabilities, and \framework{} effectively improves them. In addition, by outperforming all the ablation settings, we demonstrate that our constrained optimization algorithm proposed in \S\ref{turn3} effectively approximates better Bayesian network's conditional probabilities.

\vpara{Applying \framework{}'s Probabilities in Decision Making.}
In this section, we evaluate the performance of applying the predicted probabilities from \framework{} in decision-making tasks. In the general decision-making setting, we similarly assume a context $C=S + U$, two possible outcomes $O_1$ and $O_2$, and the task is to decide which outcome is more likely given $C$, similar to Fig.~\ref{fig:main}. The inference process we use for \framework{} is simply taking the argmax between $\mathbb{P}(O_1|C)$ and $\mathbb{P}(O_2|C)$. We remove all instances where they cannot be mapped to any factor values(i.e., \framework{}'s decision is ``unknown''). This does not undermine our experiment setting because such removal is label-agnostic, and we can always fall back to other methods such as CoT when \framework{} predicts ``unknown''.
We compare our framework with chain-of-thought inference with self-consistency on the same instances where \framework{} predictions are not ``unknown''. We present the prompt for CoT in Appendix Fig.\ref{fig:prompt-cot}.

The results are shown in Table~\ref{tab:framework}. Our first observation is that \framework{}'s probability estimations are accurate enough for direct decision making. Our proposed method outperforms the chain-of-thought baseline with Llama-3.1 on all benchmarks and performs comparably using Llama-2 as the base model. This is partially because Llama-3.1 generates more diverse and accurate factors, and we analyze this in \S\ref{appendix:error}. At the same time, we see that \framework{} performs better on the most challenging \datasetname{} dataset on both models. This demonstrates that \framework{} is a desirable solution for more complex scenarios where CoT tends to fail more. 
We also observe the inferior performance of our method compared to CoT may be attributed to the fundamental differences between probability estimation and hard label classification. Table~\ref{tb:llm_example} in the appendix shows an example that CoT sometimes does correctly due to the spurious signals and biases in the conditions, which highlights the importance of our controllable inference approach. We can conclude that \framework{} outperforms direct inference methods like CoT in most cases and is at least comparable in decision making while offering significantly better interpretability and controllability throughout the process.

\begin{table}[t]
\centering
\small
{
\begin{tabular}{lcccccc}
\toprule
Dataset & \framework{}\_1/2 & \framework{}\_1/n & \framework{}\_fixed\_prob &  \framework{}\_optimized (ours) & CoT \\
\midrule
\multicolumn{6}{c}{\textit{Llama-2-70b-instruct}} \\
\midrule
\datasetname{} &73.7&72.8 & 73.4& \textbf{73.9}& 71.5\\
\plasma{} & 72.8 & 72.3& 72.7& 74.0 & \textbf{76.8}\\
\commonsense{} &86.9&86.8 &87.5 & 89.0 & \textbf{93.8}\\
\midrule
\multicolumn{6}{c}{\textit{Llama-3.1-70b-Instruct}} \\
\midrule
\datasetname{} &65.5 & 68.9& 65.5 & \textbf{74.3}& 72.6\\
\plasma{} & 71.3 &66.5 &65.7 & \textbf{73.0} & 71.5\\
\commonsense{} &78.1&85.4&86.7& \textbf{92.3} & 90.8\\

\bottomrule
\end{tabular}
}
\caption{
Performance comparisons of \framework{} and baselines on decision-making benchmarks.
\framework{}\_optimized is our final model; chain-of-thought with self-consistency (CoT) is our main baseline; others are ablation baselines.
} 
\label{tab:framework}
\end{table}

\vpara{Ablation Studies on \framework{}'s Components.} We conduct an ablation study regarding factor generation. We compare the final decision-making accuracy based on our proposed abductive sampling method and the direct factor generation method mentioned in \S\ref{turn2}. Appendix Table~\ref{tab:ablation} shows the results: we see that our proposed method, on average, surpasses the direct method by 4.4\%/3.9\% in accuracy for Llama-2/ Llama-3.1, indicating the effectiveness of our proposed factor generation that collects more comprehensive signals from models' parametric knowledge and memorization.
We also conduct ablation studies on different ways of estimating the conditional probabilities of our Bayesian network (i.e., 1/2, 1/n, and fixed initial prob). Results in Tables~\ref{tab:main} and~\ref{tab:framework} show that our proposed constrained optimization learning algorithm effectively finds more accurate estimations by outperforming all ablation baselines. 
We demonstrate that both losses in Eq.\ref{eq:loss} are crucial in \S\ref{appendix:loss}.

\subsection{Extrinsic Applications of the \framework{} Framework}
\label{extrinsic}

\vpara{The Usage of the Reliably Estimated Probability.} 
As shown in \S\ref{sec:intrinsic-eval}, \framework{} can produce reliable probabilities that contribute to better decision making. Such findings motivate us to investigate whether these probabilities can be used as supervision signals, as models may benefit from seeing more than the discrete classification labels~\citep{zhou2020temporal}.
Specifically, we use \framework{} to estimate probabilities of some decision-making training instances and finetune a T5-large model using these generated probabilities. We then evaluate its performance on cross-domain datasets. We fully replicate the experimental settings described in~\citet{feng-etal-2023-generic} and the details can be referred to in \S\ref{appendix:probabilitytrain}. We use instances from \commonsense{}, where the argmax of the estimated probabilities from our framework using Llama-2 supports the correct outcome, as additional training data, besides the originally proposed training data in that paper. We test on two general temporal reasoning datasets, \matres{}~\citep{ning-etal-2018-multi} and \tracie{}~\citep{zhou-etal-2021-temporal}, as well as two commonsense reasoning and planning datasets, \datasetname{}, and \plasma{}. The results are shown in Table~\ref{tab:finetune}. We see that models achieve an average performance increase of 1.3\% compared with \framework{}-generated probabilities as additional signals. We use cross-entropy loss with class probabilities as the target for our soft probability setting and cross-entropy loss with class labels as the target for the hard-label setting.
Such results suggest another effective application of the \framework{} framework, where smaller models can benefit from soft probability alignment and make better overall predictions.  

\begin{table}[h]
\centering
\small
{
\begin{tabular}{lcccccc}
\toprule
Model (Train Data)&\datasetname{} (exp) &	\datasetname{}&	\tracie{}&	\matres{}&	\plasma{}	& Average \\
\cmidrule(lr){1-1}\cmidrule(lr){2-2}\cmidrule(lr){3-3}\cmidrule(lr){4-4}\cmidrule(lr){5-5}\cmidrule(lr){6-6}\cmidrule(lr){7-7}

T5\_large & 77.8 & 57.9 & \textbf{73.0} &	\textbf{83.5} &  48.1 &  65.6\\
+ hard label &	83.6 & 	62.9& 	68.2& 		75.2& \textbf{65.2} 	& 67.9	\\
+ \framework{} prob (\textit{ours})	& \textbf{84.3} & \textbf{63.4} & 71.4 &	77.6 &63.3  &  \textbf{68.9} \\
\cmidrule(lr){1-7}
PatternTime & 	82.2& 	60.9& 	\textbf{79.8}& \textbf{85.8}& 50.5	& 69.3	 \\
+ hard label	& 83.9& 	61.8& 	76.2& 	83.7& 56.9& 	69.7\\
+ \framework{} prob (\textit{ours})	&\textbf{85.1}& \textbf{62.6} & 	75.7& 	85.1& \textbf{61.5}& \textbf{71.2} \\

\bottomrule
\end{tabular}
}
\caption{System performances under different supervision data across three binary temporal
benchmarks and one binary planning benchmark. For simplicity, we use ``hard label'' representing that we use \commonsense{} supervision data with explicit binary labels, and ``\framework{} prob'' representing that we use \commonsense{} supervision data with estimated probabilities. \datasetname{} (exp) uses gold explanations during evaluation.} 
\label{tab:finetune}
\end{table}

\vpara{Using \framework{} to Generate Better Follow-up Questions.} 
Generating accurate and informative follow-up questions is an important goal for LLMs to be applied in mission-critical tasks that require unambiguous decision making. A natural benefit of our \framework{} framework is that it analyzes what needs to be known through the comprehensive and unbiased abduction step (i.e., factor generation) and interprets how new information will change the probability or confidence of a desired decision during the deduction step (i.e., probability estimation). We conduct a follow-up question generation experiment to demonstrate this benefit. 
For a given decision-making problem, we ask both \framework{} and LLMs to generate binary (yes/no) follow-up questions. For \framework{}, 
we identify implied factors, choose the most important unmentioned one, adopt the most important value aligning with the current outcome, and rephrase it as a question. We directly ask LLM to generate questions for better decision making. The details can be referred to in \S\ref{appendix:yesorno}. We then recruit crowd-workers on MTurk to provide pairwise preference between the \framework{} generated question and the question generated directly by the LLM. We do this on 250 examples from \commonsense{}. 
An example can be seen in Appendix Fig.~\ref{fig:human_annotation_1} and the details can also be seen in \S\ref{appendix:yesorno}. Appendix Table~\ref{tab:question} shows that the follow-up questions generated by \framework{} are more often preferred: questions generated by \framework{} are preferred 52.8\% of the time, while those directly generated by LLM are preferred 32.8\% of the time; in the remaining 14.4\%, both are considered equally useful. This result motivates future work on applying \framework{} in an interactive agent system for more efficient and controlled decision making.

\section{Conclusion}
In this paper, we propose \framework{}, a Bayesian inference framework for large language models. \framework{} utilizes LLMs' existing reliable capabilities such as abductive thinking and coarse-grained classification to optimize an explicit Bayesian network for more reliable probability estimation. We show that \framework{} can output accurate probabilities, outperforming LLM vanilla probabilties by 30\% on direct estimation tasks, and outperform chain-of-thought on decision-making tasks. At the same time, \framework{}'s outputs are explainable, because of its natural-language-based variables in the fully-interpretable Bayesian network. We also show that \framework{} can be used in several other settings, such as generating fine-grained supervision signals and asking informative follow-up questions, to further improve the trustworthiness and controllability of LLMs when applied to real-world applications. 
\section{Acknowledgements}
This work was sponsored by the Army Research Office and was accomplished under Grant Number W911NF-20-1-0080. The views and conclusions contained in this document are those of the authors and should not be interpreted as representing the official policies, either expressed or implied, of the Army Research Office or the U.S. Government. The U.S. Government is authorized to reproduce and distribute reprints for Government purposes notwithstanding any copyright notation herein. This work was also supported by the Office of Naval Research [ONR N00014-23-1-2364]. We thank Kyurae Kim, Ollie Liu, Vivek Gupta and colleagues at CogComp UPenn for valuable discussions.

\bibliography{main.bbl}

\begin{thebibliography}{40}
\providecommand{\natexlab}[1]{#1}
\providecommand{\url}[1]{\texttt{#1}}
\expandafter\ifx\csname urlstyle\endcsname\relax
  \providecommand{\doi}[1]{doi: #1}\else
  \providecommand{\doi}{doi: \begingroup \urlstyle{rm}\Url}\fi

\bibitem[Bordley(1982)]{bordley1982multiplicative}
Robert~F Bordley.
\newblock A multiplicative formula for aggregating probability assessments.
\newblock \emph{Management science}, 28\penalty0 (10):\penalty0 1137--1148, 1982.

\bibitem[Brahman et~al.(2023)Brahman, Bhagavatula, Pyatkin, Hwang, Li, Arai, Sanyal, Sakaguchi, Ren, and Choi]{brahman2023plasma}
Faeze Brahman, Chandra Bhagavatula, Valentina Pyatkin, Jena~D. Hwang, Xiang~Lorraine Li, Hirona~J. Arai, Soumya Sanyal, Keisuke Sakaguchi, Xiang Ren, and Yejin Choi.
\newblock Plasma: Making small language models better procedural knowledge models for (counterfactual) planning, 2023.

\bibitem[Chen et~al.(2020)Chen, Jiang, Poliak, Sakaguchi, and Van~Durme]{chen-etal-2020-uncertain}
Tongfei Chen, Zhengping Jiang, Adam Poliak, Keisuke Sakaguchi, and Benjamin Van~Durme.
\newblock Uncertain natural language inference.
\newblock In Dan Jurafsky, Joyce Chai, Natalie Schluter, and Joel Tetreault (eds.), \emph{Proceedings of the 58th Annual Meeting of the Association for Computational Linguistics}, pp.\  8772--8779, Online, July 2020. Association for Computational Linguistics.
\newblock \doi{10.18653/v1/2020.acl-main.774}.
\newblock URL \url{https://aclanthology.org/2020.acl-main.774}.

\bibitem[Feng et~al.(2023)Feng, Zhou, Wang, Jin, and Roth]{feng-etal-2023-generic}
Yu~Feng, Ben Zhou, Haoyu Wang, Helen Jin, and Dan Roth.
\newblock Generic temporal reasoning with differential analysis and explanation.
\newblock In Anna Rogers, Jordan Boyd-Graber, and Naoaki Okazaki (eds.), \emph{Proceedings of the 61st Annual Meeting of the Association for Computational Linguistics (Volume 1: Long Papers)}, pp.\  12013--12029, Toronto, Canada, July 2023. Association for Computational Linguistics.
\newblock \doi{10.18653/v1/2023.acl-long.671}.
\newblock URL \url{https://aclanthology.org/2023.acl-long.671}.

\bibitem[Friedman et~al.(1997)Friedman, Geiger, and Goldszmidt]{friedman1997bayesian}
Nir Friedman, Dan Geiger, and Moises Goldszmidt.
\newblock Bayesian network classifiers.
\newblock \emph{Machine learning}, 29:\penalty0 131--163, 1997.

\bibitem[Gelman et~al.(1995)Gelman, Carlin, Stern, and Rubin]{reason:GelCarSteRub95}
Andrew~B. Gelman, John~B. Carlin, Hal~S. Stern, and Donald~B. Rubin.
\newblock \emph{Bayesian Data Analysis}.
\newblock Chapman and Hall/CRC, Boca Ratan, Florida, 1995.

\bibitem[Han et~al.(2024)Han, Buntine, and Shareghi]{han2024uncertaintyaware}
Jiuzhou Han, Wray Buntine, and Ehsan Shareghi.
\newblock Towards uncertainty-aware language agent, 2024.

\bibitem[Hoffman et~al.(2023)Hoffman, Phan, david dohan, Douglas, Le, Parisi, Sountsov, Sutton, Vikram, and Saurous]{hoffman2023training}
Matthew~Douglas Hoffman, Du~Phan, david dohan, Sholto Douglas, Tuan~Anh Le, Aaron~T Parisi, Pavel Sountsov, Charles Sutton, Sharad Vikram, and Rif~A. Saurous.
\newblock Training chain-of-thought via latent-variable inference.
\newblock In \emph{Thirty-seventh Conference on Neural Information Processing Systems}, 2023.
\newblock URL \url{https://openreview.net/forum?id=a147pIS2Co}.

\bibitem[Hou et~al.(2024)Hou, Liu, Qian, Andreas, Chang, and Zhang]{pmlr-v235-hou24b}
Bairu Hou, Yujian Liu, Kaizhi Qian, Jacob Andreas, Shiyu Chang, and Yang Zhang.
\newblock Decomposing uncertainty for large language models through input clarification ensembling.
\newblock In Ruslan Salakhutdinov, Zico Kolter, Katherine Heller, Adrian Weller, Nuria Oliver, Jonathan Scarlett, and Felix Berkenkamp (eds.), \emph{Proceedings of the 41st International Conference on Machine Learning}, volume 235 of \emph{Proceedings of Machine Learning Research}, pp.\  19023--19042. PMLR, 21--27 Jul 2024.
\newblock URL \url{https://proceedings.mlr.press/v235/hou24b.html}.

\bibitem[Hu et~al.(2024)Hu, Jain, Elmoznino, Kaddar, Lajoie, Bengio, and Malkin]{hu2024amortizing}
Edward~J Hu, Moksh Jain, Eric Elmoznino, Younesse Kaddar, Guillaume Lajoie, Yoshua Bengio, and Nikolay Malkin.
\newblock Amortizing intractable inference in large language models.
\newblock In \emph{The Twelfth International Conference on Learning Representations}, 2024.
\newblock URL \url{https://openreview.net/forum?id=Ouj6p4ca60}.

\bibitem[Kahneman(2003)]{psy_1}
Daniel Kahneman.
\newblock A perspective on judgment and choice: mapping bounded rationality.
\newblock In \emph{The American psychologist vol. 58,9 (2003): 697-720}, 2003.
\newblock \doi{10.1037/0003-066X.58.9.697}.

\bibitem[Kambhampati et~al.(2024)Kambhampati, Valmeekam, Guan, Verma, Stechly, Bhambri, Saldyt, and Murthy]{kambhampati2024position}
Subbarao Kambhampati, Karthik Valmeekam, Lin Guan, Mudit Verma, Kaya Stechly, Siddhant Bhambri, Lucas~Paul Saldyt, and Anil~B Murthy.
\newblock Position: {LLM}s can{\textquoteright}t plan, but can help planning in {LLM}-modulo frameworks.
\newblock In \emph{Forty-first International Conference on Machine Learning}, 2024.
\newblock URL \url{https://openreview.net/forum?id=Th8JPEmH4z}.

\bibitem[Kassner et~al.(2023)Kassner, Tafjord, Sabharwal, Richardson, Schuetze, and Clark]{kassner-etal-2023-language}
Nora Kassner, Oyvind Tafjord, Ashish Sabharwal, Kyle Richardson, Hinrich Schuetze, and Peter Clark.
\newblock Language models with rationality.
\newblock In Houda Bouamor, Juan Pino, and Kalika Bali (eds.), \emph{Proceedings of the 2023 Conference on Empirical Methods in Natural Language Processing}, pp.\  14190--14201, Singapore, December 2023. Association for Computational Linguistics.
\newblock \doi{10.18653/v1/2023.emnlp-main.877}.
\newblock URL \url{https://aclanthology.org/2023.emnlp-main.877}.

\bibitem[Koller \& Friedman(2009)Koller and Friedman]{10.5555/1795555}
Daphne Koller and Nir Friedman.
\newblock \emph{Probabilistic Graphical Models: Principles and Techniques - Adaptive Computation and Machine Learning}.
\newblock The MIT Press, 2009.
\newblock ISBN 0262013193.

\bibitem[Kuhn et~al.(2023)Kuhn, Gal, and Farquhar]{kuhn2023semantic}
Lorenz Kuhn, Yarin Gal, and Sebastian Farquhar.
\newblock Semantic uncertainty: Linguistic invariances for uncertainty estimation in natural language generation.
\newblock In \emph{The Eleventh International Conference on Learning Representations}, 2023.
\newblock URL \url{https://openreview.net/forum?id=VD-AYtP0dve}.

\bibitem[Lew et~al.(2020)Lew, Tessler, Mansinghka, and Tenenbaum]{lew2020leveraging}
Alexander~K Lew, Michael~Henry Tessler, Vikash~K Mansinghka, and Joshua~B Tenenbaum.
\newblock Leveraging unstructured statistical knowledge in a probabilistic language of thought.
\newblock In \emph{Proceedings of the annual conference of the cognitive science society}, 2020.

\bibitem[Li et~al.(2024)Li, Zhou, Wang, Fu, Roth, and Chen]{li-etal-2024-deceptive}
Bangzheng Li, Ben Zhou, Fei Wang, Xingyu Fu, Dan Roth, and Muhao Chen.
\newblock Deceptive semantic shortcuts on reasoning chains: How far can models go without hallucination?
\newblock In Kevin Duh, Helena Gomez, and Steven Bethard (eds.), \emph{Proceedings of the 2024 Conference of the North American Chapter of the Association for Computational Linguistics: Human Language Technologies (Volume 1: Long Papers)}, pp.\  7675--7688, Mexico City, Mexico, June 2024. Association for Computational Linguistics.
\newblock \doi{10.18653/v1/2024.naacl-long.424}.
\newblock URL \url{https://aclanthology.org/2024.naacl-long.424}.

\bibitem[Lin et~al.(2023)Lin, Fu, Yang, Ammanabrolu, Brahman, Huang, Bhagavatula, Choi, and Ren]{Lin2023SwiftSageAG}
Bill~Yuchen Lin, Yicheng Fu, Karina Yang, Prithviraj Ammanabrolu, Faeze Brahman, Shiyu Huang, Chandra Bhagavatula, Yejin Choi, and Xiang Ren.
\newblock Swiftsage: A generative agent with fast and slow thinking for complex interactive tasks.
\newblock \emph{ArXiv}, abs/2305.17390, 2023.

\bibitem[Lin et~al.(2024)Lin, Trivedi, and Sun]{lin2024generatingconfidenceuncertaintyquantification}
Zhen Lin, Shubhendu Trivedi, and Jimeng Sun.
\newblock Generating with confidence: Uncertainty quantification for black-box large language models, 2024.
\newblock URL \url{https://arxiv.org/abs/2305.19187}.

\bibitem[Liu et~al.(2025)Liu, Fu, Yogatama, and Neiswanger]{liu2025dellma}
Ollie Liu, Deqing Fu, Dani Yogatama, and Willie Neiswanger.
\newblock De{LLM}a: Decision making under uncertainty with large language models.
\newblock In \emph{The Thirteenth International Conference on Learning Representations}, 2025.
\newblock URL \url{https://openreview.net/forum?id=Acvo2RGSCy}.

\bibitem[McCarthy \& Hayes(1981)McCarthy and Hayes]{mccarthy1981some}
John McCarthy and Patrick~J Hayes.
\newblock Some philosophical problems from the standpoint of artificial intelligence.
\newblock In \emph{Readings in artificial intelligence}, pp.\  431--450. Elsevier, 1981.

\bibitem[McGlynn(2014)]{McGlynn2014ThinkingFA}
N.~F. McGlynn.
\newblock Thinking fast and slow.
\newblock \emph{Australian veterinary journal}, 92 12:\penalty0 N21, 2014.

\bibitem[Ning et~al.(2018)Ning, Wu, and Roth]{ning-etal-2018-multi}
Qiang Ning, Hao Wu, and Dan Roth.
\newblock A multi-axis annotation scheme for event temporal relations.
\newblock In \emph{Proceedings of the 56th Annual Meeting of the Association for Computational Linguistics (Volume 1: Long Papers)}, pp.\  1318--1328, Melbourne, Australia, July 2018. Association for Computational Linguistics.
\newblock \doi{10.18653/v1/P18-1122}.
\newblock URL \url{https://aclanthology.org/P18-1122}.

\bibitem[Ozturkler et~al.(2023)Ozturkler, Malkin, Wang, and Jojic]{ozturkler-etal-2023-thinksum}
Batu Ozturkler, Nikolay Malkin, Zhen Wang, and Nebojsa Jojic.
\newblock {T}hink{S}um: Probabilistic reasoning over sets using large language models.
\newblock In Anna Rogers, Jordan Boyd-Graber, and Naoaki Okazaki (eds.), \emph{Proceedings of the 61st Annual Meeting of the Association for Computational Linguistics (Volume 1: Long Papers)}, Toronto, Canada, July 2023. Association for Computational Linguistics.
\newblock \doi{10.18653/v1/2023.acl-long.68}.
\newblock URL \url{https://aclanthology.org/2023.acl-long.68}.

\bibitem[Ren et~al.(2023)Ren, Dixit, Bodrova, Singh, Tu, Brown, Xu, Takayama, Xia, Varley, Xu, Sadigh, Zeng, and Majumdar]{ren2023robots}
Allen~Z. Ren, Anushri Dixit, Alexandra Bodrova, Sumeet Singh, Stephen Tu, Noah Brown, Peng Xu, Leila Takayama, Fei Xia, Jake Varley, Zhenjia Xu, Dorsa Sadigh, Andy Zeng, and Anirudha Majumdar.
\newblock Robots that ask for help: Uncertainty alignment for large language model planners.
\newblock In \emph{7th Annual Conference on Robot Learning}, 2023.
\newblock URL \url{https://openreview.net/forum?id=4ZK8ODNyFXx}.

\bibitem[Roth \& Yih(2005)Roth and Yih]{roth2005integer}
Dan Roth and Wen-tau Yih.
\newblock Integer linear programming inference for conditional random fields.
\newblock In \emph{Proceedings of the 22nd international conference on Machine learning}, pp.\  736--743, 2005.

\bibitem[Singh et~al.(2021)Singh, Wen, Hou, Alipoormolabashi, Wu, Ma, and Peng]{singh-etal-2021-com2sense}
Shikhar Singh, Nuan Wen, Yu~Hou, Pegah Alipoormolabashi, Te-lin Wu, Xuezhe Ma, and Nanyun Peng.
\newblock {COM}2{SENSE}: A commonsense reasoning benchmark with complementary sentences.
\newblock In \emph{Findings of the Association for Computational Linguistics: ACL-IJCNLP 2021}, pp.\  883--898, Online, August 2021. Association for Computational Linguistics.
\newblock \doi{10.18653/v1/2021.findings-acl.78}.
\newblock URL \url{https://aclanthology.org/2021.findings-acl.78}.

\bibitem[Tafjord et~al.(2022)Tafjord, Dalvi~Mishra, and Clark]{tafjord-etal-2022-entailer}
Oyvind Tafjord, Bhavana Dalvi~Mishra, and Peter Clark.
\newblock Entailer: Answering questions with faithful and truthful chains of reasoning.
\newblock In Yoav Goldberg, Zornitsa Kozareva, and Yue Zhang (eds.), \emph{Proceedings of the 2022 Conference on Empirical Methods in Natural Language Processing}, pp.\  2078--2093, Abu Dhabi, United Arab Emirates, December 2022. Association for Computational Linguistics.
\newblock \doi{10.18653/v1/2022.emnlp-main.134}.
\newblock URL \url{https://aclanthology.org/2022.emnlp-main.134}.

\bibitem[Tian et~al.(2023)Tian, Mitchell, Zhou, Sharma, Rafailov, Yao, Finn, and Manning]{tian-etal-2023-just}
Katherine Tian, Eric Mitchell, Allan Zhou, Archit Sharma, Rafael Rafailov, Huaxiu Yao, Chelsea Finn, and Christopher Manning.
\newblock Just ask for calibration: Strategies for eliciting calibrated confidence scores from language models fine-tuned with human feedback.
\newblock In Houda Bouamor, Juan Pino, and Kalika Bali (eds.), \emph{Proceedings of the 2023 Conference on Empirical Methods in Natural Language Processing}, pp.\  5433--5442, Singapore, December 2023. Association for Computational Linguistics.
\newblock \doi{10.18653/v1/2023.emnlp-main.330}.
\newblock URL \url{https://aclanthology.org/2023.emnlp-main.330}.

\bibitem[TVERSKY \& KAHNEMAN(1974)TVERSKY and KAHNEMAN]{psy_2}
AMOS TVERSKY and DANIEL KAHNEMAN.
\newblock Judgment under uncertainty: Heuristics and biases.
\newblock In \emph{Science, 185(4157), 1124–1131}, 1974.
\newblock URL \url{https://doi.org/10.1126/science.185.4157.1124}.

\bibitem[Wang et~al.(2023)Wang, Wei, Schuurmans, Le, Chi, Narang, Chowdhery, and Zhou]{wang2023selfconsistency}
Xuezhi Wang, Jason Wei, Dale Schuurmans, Quoc~V Le, Ed~H. Chi, Sharan Narang, Aakanksha Chowdhery, and Denny Zhou.
\newblock Self-consistency improves chain of thought reasoning in language models.
\newblock In \emph{The Eleventh International Conference on Learning Representations}, 2023.
\newblock URL \url{https://openreview.net/forum?id=1PL1NIMMrw}.

\bibitem[Wei et~al.(2022)Wei, Wang, Schuurmans, Bosma, brian ichter, Xia, Chi, Le, and Zhou]{wei2022chain}
Jason Wei, Xuezhi Wang, Dale Schuurmans, Maarten Bosma, brian ichter, Fei Xia, Ed~H. Chi, Quoc~V Le, and Denny Zhou.
\newblock Chain of thought prompting elicits reasoning in large language models.
\newblock In Alice~H. Oh, Alekh Agarwal, Danielle Belgrave, and Kyunghyun Cho (eds.), \emph{Advances in Neural Information Processing Systems}, 2022.
\newblock URL \url{https://openreview.net/forum?id=_VjQlMeSB_J}.

\bibitem[Wolfson et~al.(2020)Wolfson, Geva, Gupta, Gardner, Goldberg, Deutch, and Berant]{wolfson-etal-2020-break}
Tomer Wolfson, Mor Geva, Ankit Gupta, Matt Gardner, Yoav Goldberg, Daniel Deutch, and Jonathan Berant.
\newblock Break it down: A question understanding benchmark.
\newblock \emph{Transactions of the Association for Computational Linguistics}, 8:\penalty0 183--198, 2020.
\newblock \doi{10.1162/tacl_a_00309}.
\newblock URL \url{https://aclanthology.org/2020.tacl-1.13}.

\bibitem[Wong et~al.(2023)Wong, Grand, Lew, Goodman, Mansinghka, Andreas, and Tenenbaum]{wong2023wordmodelsworldmodels}
Lionel Wong, Gabriel Grand, Alexander~K. Lew, Noah~D. Goodman, Vikash~K. Mansinghka, Jacob Andreas, and Joshua~B. Tenenbaum.
\newblock From word models to world models: Translating from natural language to the probabilistic language of thought, 2023.
\newblock URL \url{https://arxiv.org/abs/2306.12672}.

\bibitem[Xiong et~al.(2024)Xiong, Hu, Lu, LI, Fu, He, and Hooi]{xiong2024can}
Miao Xiong, Zhiyuan Hu, Xinyang Lu, YIFEI LI, Jie Fu, Junxian He, and Bryan Hooi.
\newblock Can {LLM}s express their uncertainty? an empirical evaluation of confidence elicitation in {LLM}s.
\newblock In \emph{The Twelfth International Conference on Learning Representations}, 2024.
\newblock URL \url{https://openreview.net/forum?id=gjeQKFxFpZ}.

\bibitem[Yadkori et~al.(2024)Yadkori, Kuzborskij, György, and Szepesvári]{yadkori2024believebelievellm}
Yasin~Abbasi Yadkori, Ilja Kuzborskij, András György, and Csaba Szepesvári.
\newblock To believe or not to believe your llm, 2024.
\newblock URL \url{https://arxiv.org/abs/2406.02543}.

\bibitem[Zhou et~al.(2020)Zhou, Ning, Khashabi, and Roth]{zhou2020temporal}
Ben Zhou, Qiang Ning, Daniel Khashabi, and Dan Roth.
\newblock Temporal common sense acquisition with minimal supervision.
\newblock In \emph{Proceedings of the 58th Annual Meeting of the Association for Computational Linguistics}, pp.\  7579--7589, 2020.

\bibitem[Zhou et~al.(2021)Zhou, Richardson, Ning, Khot, Sabharwal, and Roth]{zhou-etal-2021-temporal}
Ben Zhou, Kyle Richardson, Qiang Ning, Tushar Khot, Ashish Sabharwal, and Dan Roth.
\newblock Temporal reasoning on implicit events from distant supervision.
\newblock In \emph{Proceedings of the 2021 Conference of the North American Chapter of the Association for Computational Linguistics: Human Language Technologies}, pp.\  1361--1371, Online, June 2021. Association for Computational Linguistics.
\newblock \doi{10.18653/v1/2021.naacl-main.107}.
\newblock URL \url{https://aclanthology.org/2021.naacl-main.107}.

\bibitem[Zhou et~al.(2022)Zhou, Richardson, Yu, and Roth]{zhou-etal-2022-learning-decompose}
Ben Zhou, Kyle Richardson, Xiaodong Yu, and Dan Roth.
\newblock Learning to decompose: Hypothetical question decomposition based on comparable texts.
\newblock In Yoav Goldberg, Zornitsa Kozareva, and Yue Zhang (eds.), \emph{Proceedings of the 2022 Conference on Empirical Methods in Natural Language Processing}, pp.\  2223--2235, Abu Dhabi, United Arab Emirates, December 2022. Association for Computational Linguistics.
\newblock \doi{10.18653/v1/2022.emnlp-main.142}.
\newblock URL \url{https://aclanthology.org/2022.emnlp-main.142}.

\bibitem[Zhou et~al.(2024)Zhou, Zhang, Chen, Yu, Wang, Peng, Roth, and Yu]{zhou2024conceptual}
Ben Zhou, Hongming Zhang, Sihao Chen, Dian Yu, Hongwei Wang, Baolin Peng, Dan Roth, and Dong Yu.
\newblock Conceptual and unbiased reasoning in language models.
\newblock \emph{arXiv preprint arXiv:2404.00205}, 2024.

\end{thebibliography}
\bibliographystyle{iclr2025_conference}

\clearpage
\appendix
\appendix
\section{Appendix}

\subsection{Derivation of the equation for uncertainty estimation}
\label{appendix:derivation}
We derive the outcome probability for the given context $C$ of partial information, i.e., any additional condition $U$ under the scenario $S$, based on the complete information space $ \mathcal{F}$ of scenario $S$:

\begin{align}
\begin{split}
\mathbb{P}(O_i|C) &= \sum_{f\in \mathcal{F}} \mathbb{P}(O_i|f, C)\mathbb{P}(f|C) \\
&= \sum_{f\in \mathcal{F}} \mathbb{P}(O_i|f, C)\mathbb{P}(f|C) (\mathbbm{1}\{f \cap C = f\} + \mathbbm{1}\{f \cap C \neq f\})\\
&= \sum_{f\in \mathcal{F} \cap \{f \cap C = f\}} \mathbb{P}(O_i|f, C)\mathbb{P}(f|C) + \sum_{f\in \mathcal{F} \cap \{f \cap C \neq f\}} \mathbb{P}(O_i|f, C)\mathbb{P}(f|C) \\
&= \sum_{f\in \mathcal{F} \cap \{ f \cap C = f\}} \mathbb{P}(O_i|f)\mathbb{P}(f|C) + \sum_{f\in \mathcal{F} \cap \{ f \cap C \neq f\}} \mathbb{P}(O_i|f, C)\mathbb{P}(f|C) \\
&= \sum_{f\in \mathcal{F} \cap \{ f \cap C = f\}} \mathbb{P}(O_i|f)\mathbb{P}(f|C) + \sum_{f\in \mathcal{F} \cap \{ f \cap C \neq f\}} \mathbb{P}(O_i|f)\mathbb{P}(f|C) \\
&= \sum_{f\in \mathcal{F}} \mathbb{P}(O_i|f)\mathbb{P}(f|C)
\end{split}
\end{align}

Where the fourth line follows from that $\mathbb{P}(O_i|f,C) = \mathbb{P}(O_i|f) $ since  $f \cap C = f$. And the second last line is derived from that $\mathbb{P}(f|C) = 0$ if $f \cap C \neq f$  as for a given $C$, only one value can be chosen for each factor as in Fig.~\ref{fig:main}. 

\subsection{Assumptions for estimating $\mathbb{P}(O_i|f)$}
\label{appendix:assumption} 

\vpara{Weak ordering of the set of factor odds ratios.} Suppose given factors $f_1, f_2, ..., f_N$, we assess odds of $o_1, o_2,..., o_N$,  respectively in favor of the outcome $O_i, i =1,2$ occurring. $o_j = \frac{\mathbb{P}(O_i|f_j)}{1-\mathbb{P}(O_i|f_j)}, j= 1,...,N$, the intuitive ordering $\geq_{E}$ is defined as:

\begin{equation}
    (o_1, o_2,..., o_N) \geq_{E} (o_1^{*}, o_2^{*},..., o_N^{*})
\end{equation}

if and only if the decision maker feels at least as sure about the outcome $O_i$, occurring
when given factors $f_1, f_2, ..., f_N$ we assess odds $o_1, o_2,..., o_N$ -as when given factors $f_1, f_2, ..., f_N$ we assess odds $o_1^{*}, o_2^{*},..., o_N^{*}$. The ordering $\geq_{E}$ is a Weak Ordering if and only if for any
$(o_1, o_2,..., o_N)$,$(o_1^{*}, o_2^{*},..., o_N^{*})$,$(o_1^{**}, o_2^{**},..., o_N^{**})$, we have:
\begin{itemize}
 \item 
(a) either $(o_1, o_2,..., o_N)$ $\geq_{E}$ $(o_1^{*}, o_2^{*},..., o_N^{*})$ or $(o_1^{*}, o_2^{*},..., o_N^{*})$ $\geq_{E}$ $(o_1, o_2,..., o_N)$,
 \item 
(b) if $(o_1, o_2,..., o_N)$ $\geq_{E}$ $(o_1^{*}, o_2^{*},..., o_N^{*})$ and if $(o_1^{*}, o_2^{*},..., o_N^{*})$ $\geq_{E}$ $(o_1^{**}, o_2^{**},..., o_N^{**})$ then $(o_1, o_2,..., o_N)$ $\geq_{E}$ $(o_1^{**}, o_2^{**},..., o_N^{**})$.
\end{itemize}

\vpara{``Noninteraction'' property of factors.} The non-interaction condition says that the decision maker's interpretation of how one factor affects the outcome does not depend on how other factors affect the outcome.

\subsection{Learning algorithm for estimating $\mathbb{P}(O_i|f)$}
\label{appendix:algorithm}
We present the complete algorithm in Appendix Algorithm~\ref{tab: algorithm}. Since $\mathbb{P}(O_1)$ and $\mathbb{P}(O_2)$ are complementary, we only need to learn $\mathbb{P}(O_1|f_{ja}),f_{ja} \in \mathcal{F}_j, j = 1, ..., N, a =1, ..., {\rm card}(\mathcal{F}_j)$, and $\mathbb{P}(O_2|f_{ja}) = 1 - \mathbb{P}(O_1|f_{ja})$. As a result, for a given scenario $S$, the number of learnable parameters is the total number of values for all the generated factors, i.e., $\sum_{j=1}^{N}{\rm card}(\mathcal{F}_j)$. 

All training instances for a scenario are drawn from the complete information space for the same scenario $S$  and possible outcomes $\{O\}$ and will share the same factors  $\{F_j\}_{j=1}^N$. We randomly sample 128 instances in the space of complete information as the training data for each scenario. This represents the minimum number of instances required for effective training. Ideally, increasing the number of sampled instances would improve the model's alignment with the underlying LLM. For a scenario with fewer than 128 instances in the space of complete information, we employ a resampling technique, selecting instances repeatedly until we reach a total of 128 samples. 

The value of $\mathbb{P}_{\rm init}(O_i|f_j)$ in Eq.\ref{eq:initial_prob} is assigned as it represents random initialization.
To determine the mappings of rankings to probabilities in the algorithm, we consulted two psychology experts and adopted the Likert scale theory. The finalized mapping between verbalized probabilities and numerical probabilities reflects typical human behavior and effectively distinguishes between verbalized probabilities in an unbiased manner. Additionally, slight adjustments to the mappings do not noticeably affect overall performance. All training hyperparameters for the algorithm are optimized using a grid search with a hold-out validation set from \plasma{}.

\begin{algorithm}[]
\DontPrintSemicolon
\SetAlgoLined
\SetKwInOut{Input}{input}
\SetKwInOut{Output}{output}
\SetKwComment{Comment}{\textcolor{blue}{\# }}{}
\caption{\label{tab: algorithm}Training single probabilities to estimate $\mathbb{P}(O_i|f)$ .}
\Input{Training set of complete information \( \{f\}, f \in \mathcal{F} \), outcome \(O_i, i=1,2 \), LLM \( M\), learning rate \( \eta \), number of epochs \( E \), batch size \( B \).}
\Output{Trained probabilities  \( \mathbb{P}(O_i|f_{ja}), f_{ja} \in \mathcal{F}_j, j = 1, ..., N, a =1, ..., {\rm card}(\mathcal{F}_j), i = 1,2\).}
\BlankLine
\BlankLine \Comment*[l]{\textcolor{blue}{Get verbalized probability scores from the LLM.}}
\For{\( f^{'} \in \{f\}\)}{
\(\mathbb{P}_{\rm LLM}(O_1|f^{'}), \mathbb{P}_{\rm LLM}(O_2|f^{'}) \leftarrow M(f^{'},O_1,O_2) \)}
\BlankLine
\BlankLine  \Comment*[l]{\textcolor{blue}{Initialize probabilities and calculate target for the margin ranking loss.}}
\For{\( j = 1 \) \KwTo \( N \)}{
\For{\( a = 1\) \KwTo \( {\rm card}(\mathcal{F}_j) \)}{
\(\mathbb{P}_{\rm init}(O_1|f_{ja})  = \left\{ 
    \begin{array}{ll}
       75\% & f_{ja} \  {\rm supports}  \ {\rm outcome} \ 1\\
       50\% & f_{ja} \ {\rm is} \ {\rm neutral} \\
       25\% & f_{ja} \  {\rm supports}  \ {\rm opposite} \ {\rm outcome} \ 2 \\
    \end{array}
\right.\)

\(y_{\rm target}(O_1|f_{ja}) =  \left\{ 
    \begin{array}{ll}
       1 & \mathbb{P}_{\rm init}(O_1|f_{ja})  > 50\% \\
       0 & \mathbb{P}_{\rm init}(O_1|f_{ja})  = 50\% \\
       -1 & \mathbb{P}_{\rm init}(O_1|f_{ja})  < 50\% \\
    \end{array}
\right.\)
}}

\BlankLine
\BlankLine  \Comment*[l]{\textcolor{blue}{Training.}}
\For{\( epoch = 1 \) \KwTo \( E \)}{
  \For {each batch of $B$ samples from the training set}{
  \For{\( b \in 1,..., B \)(in parallel)} {
  \(\mathbb{P}_{\rm estimated}(O_1|f^{b}) = \frac{\prod_{j=1}^N\mathbb{P}(O_1|f^{b}_j)}{\prod_{j=1}^N\mathbb{P}(O_1|f^{b}_j) + \prod_{j=1}^N(1-\mathbb{P}(O_1|f^{b}_j))}\)
 \Comment*[l]{\textcolor{blue}{Forward propagation.}}

   \(\ell_{\rm MSE}^b = (\mathbb{P}_{\rm LLM}(O_1|f^{b})- \mathbb{P}_{\rm estimated}(O_1|f^{b}))^2\) \Comment*[l]{\textcolor{blue}{Compute MSE loss for each sample, for \(f_{ja} \in \mathcal{F}_j, f_{ja} \neq f^{b}_j, \frac{\partial\ell_{\rm MSE}^b}{\partial\mathbb{P}(O_1|f_{ja})} = 0, j = 1, ..., N\).}}
  }
   \BlankLine  
  \( \ell_{\rm MSE} = \frac{1}{B} \sum_{b=1}^{B} \ell_{\rm MSE}^b  \) \Comment*[l]{\textcolor{blue}{Compute MSE loss.}}
  \BlankLine
   \For{\( j = 1 \) \KwTo \( N \)}{
   
   \For{\( a = 1\) \KwTo \( {\rm card}(\mathcal{F}_{j}) \)}{
\(\mathbb{P}_{\rm trained}(O_1|f_{ja}) = \sum_{f^{*} \in \mathcal{F}^*} \mathbb{P}_{\rm estimated}(O_1|f^{*} \cup f_{ja} ) \mathbb{P}(f^{*}  | f_{ja} ) = \frac{1}{{\rm card}(\mathcal{F}^*)}\sum_{f^{*} \in \mathcal{F}^*}\mathbb{P}_{\rm estimated}(O_1|f^{*} \cup f_{ja} )\)

   \(\ell^{ja}_{\rm MR} =  {\rm max}(0, -y_{\rm target}(O_1|f_{ja}) (\mathbb{P}_{\rm trained}(O_1|f_{ja})- 0.5) + \epsilon)\)
   }}
   \(\ell_{\rm MR} = \frac{1}{N} \sum_{j=1}^{N} \frac{1}{{\rm card}(\mathcal{F}_{j})} \sum_{a=1}^{{\rm card}(\mathcal{F}_{j})}\ell^{ja}_{\rm MR}  \)  \Comment*[l]{\textcolor{blue}{Compute MR loss.}}

  \BlankLine
  \( \mathcal{L} = \ell_{\rm MSE} + \alpha * \ell_{\rm MR} \)\Comment*[l]{\textcolor{blue}{Compute final loss.} }

   \BlankLine
  \For{\( j = 1 \) \KwTo \( N \)}{
\For{\( a = 1\) \KwTo \( {\rm card}(\mathcal{F}_j) \)}{
  \( \mathbb{P}(O_1|f_{ja}) =  \mathbb{P}(O_1|f_{ja}) - \eta \frac{\partial\mathcal{L}}{\partial\mathbb{P}(O_1|f_{ja})}  \)  \Comment*[l]{\textcolor{blue}{Update parameters.}}
  
  \( \mathbb{P}(O_2|f_{ja}) = 1- \mathbb{P}(O_1|f_{ja})\)

  }}
} 
}

\end{algorithm}

\subsection{Experiment Dataset Description}
We convert three datasets focusing on reasoning and planning to our problem setting.
\label{appendix:dataset}
\begin{itemize}
    \item \textbf{\commonsense{}}~\citep{singh-etal-2021-com2sense}: focuses on commonsense reasoning and planning. The original task constitutes a multi-domain complementary commonsense reasoning benchmark and comprises natural language true/false statements. We exclusively focus on instances of comparative reasoning where a pre-trained smaller model exhibits low confidence in its prediction logits. Using GPT-4\footnote{We use GPT-4-0613 when utilizing GPT-4 for annotations.}, we rewrite each instance statement into its opposite, creating two distinct outcomes. We then outline a general scenario based on the two statements and generate 10 additional sentences for each statement, with detailed conditions that would support the statement. Additionally, we implement a reverse verification step to eliminate any incorrect additional sentences to ensure dataset quality. We gather 216 scenarios and 3822 instances. 9 additional conditions on average support each outcome. 

    \item \textbf{\datasetname{}}~\citep{feng-etal-2023-generic}: focuses on temporal reasoning. The dataset is designed to study how introducing an extra sentence to a specific scenario affects the temporal relation, aligning well with the context of our task. There are 1000 instances. 
    
    \item \textbf{\plasma{}}~\citep{brahman2023plasma}: focuses on planning. The original task is to generate or revise a plan based on a given goal while adhering to an additional condition. We've observed that the key variations between the initial and the revised plans given the additional condition typically focus on a specific step within the plan. We then utilize GPT-4 to identify the step and adapt the description of the step in the original and revised plans into two potential outcomes. We manually identify the less common outcome in the 279 scenarios and use GPT-4 to generate 5 additional sentences for the outcome, with detailed conditions that would support this outcome instead of the other. We gather 279 scenarios and 1395 instances with 5 additional conditions for each scenario. 
\end{itemize}

For the two GPT-4 generated datasets, we recruited the authors and their colleagues to conduct a human quality check on 100 randomly selected instances from each dataset, with three people reviewing each instance. All three reviewers agreed on the label for 91\% of the generated data. For 94\% of the data, at least two reviewers agreed on the label.

 \begin{figure*}[ht]
\centering
\scalebox{1}{
\includegraphics[width=0.98\textwidth]{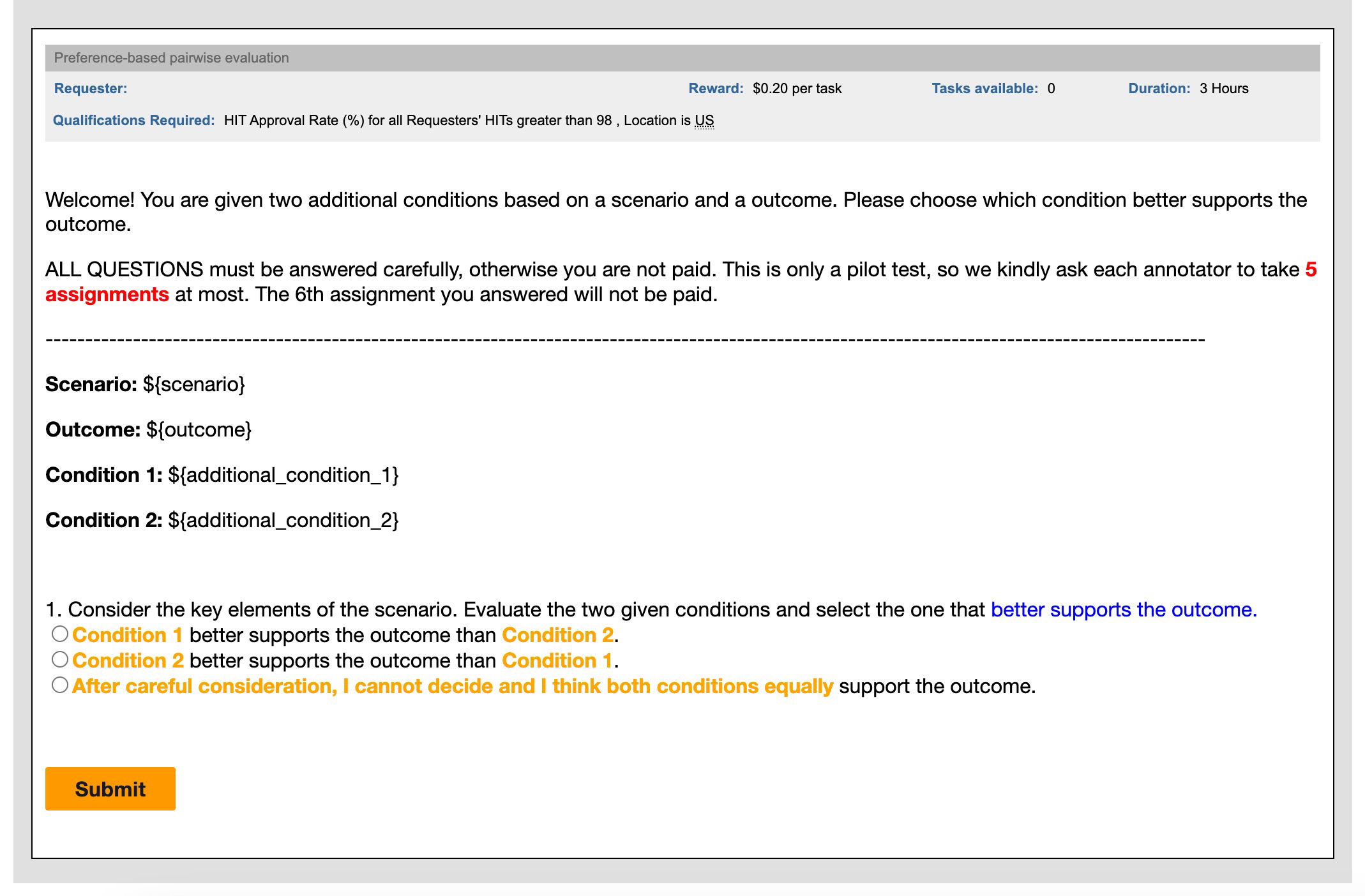}}
	\caption{\label{fig:human_annotation}The interface for human evaluation on preference-based pairwise evaluation of the estimated probabilities.}
\end{figure*}

\subsection{Human evaluation on \framework{} estimated probabilities}
\label{appendix:humaneval}
We first explain our choice of relative evaluation over asking human annotators to provide absolute probabilities from a psychological perspective. \cite{psy_2} shows that humans use heuristic principles to make probability judgments,  which often suffer from systematic and predictable errors due to cognitive biases. \cite{psy_1} further discusses how human judgment in probabilistic situations is generally better in relative than absolute terms due to cognitive biases and limitations in processing complex information. We conducted a pilot study with human annotators on MTurk, asking them to provide absolute probabilities for each additional condition separately. However, we were unable to reach a consensus among the 7 annotators.

We then introduce the human annotation process for pairwise comparison. We collect human gold labels with rigorous quality control for the preference-based pairwise evaluation of the estimated probabilities. We restrict each annotator to take 5 samples at most and there are 386 distinct annotators. We only allow participants who have 98\% or more HITs acceptance rate and are located in the US. We also require annotators to spend at least 1 minute for each instance (the hourly
salary is ~\$15). We show the experiment interfaces in Appendix Fig.~\ref{fig:human_annotation}. We first carry out the human annotation process five times on 450 randomly sampled instances from \commonsense{} when the condition can be mapped to at least one factor and therefore \framework{} does not produce ``unknown'' during our experiments, and both conditions support one target outcome. If the 5 annotations have a less than 80\% agreement, i.e., less than 4 out of the 5 annotators agree on the same condition label, we invite 2 additional annotators to reassess this instance. If the agreement of the 7 annotators is more than 70\%, i.e., at least 5 out of the 7 annotators agree on the same condition label, we accept the instance. Other than the previously accepted instances, we also accept an instance as a neutral label if all three labels have been chosen during the annotations and the agreement for the neutral label is more than or equal 40\% and the agreement for any of the remaining two labels is less than or equal 40\%. 
We get 350 instances in total, where 12.3\% of the instances have a neutral label where condition 1 and condition 2 equally support the outcome, 44.0\% of the instances have a label towards condition 1 where condition 1 supports the corresponding outcome better than condition 2, 43.7\% of the instances have a label towards condition 2 where condition 2 supports the corresponding outcome better than condition 1.

We now introduce how precision and recall are calculated. Given a testing instance, assuming the gold outcome is $O^*$ and the opposite outcome is $O^\neg$, there will be 3 model output labels: 1) $P(O^*|C1) > P(O^*|C2)$, i.e., the model thinks C1 better support the gold outcome; 2) $P(O^*|C2) > P(O^*|C1)$, i.e., the model thinks C2 better support the gold outcome; 3) $P(O^*|C2) = P(O^*|C1)$, i.e., the model thinks both $C_1$ and $C_2$ equally support the gold outcome. We want to check if the model's label is the same as the human annotator's label. Therefore, this can further be seen as a 3-label classification problem. 

\subsection{Baseline Description}
We adopt the following baselines to evaluate our \framework{} framework. 
\label{appendix:baseline}
\begin{itemize}
    \item \textbf{Vanilla Verbalization with Self-Consistency (Vanilla).}~\citep{wang2023selfconsistency} Given the scenario, an additional condition, and two corresponding outcomes, the model is asked to directly verbalize an estimated probability for each outcome. We show the Prompt in Appendix Fig.~\ref{tab:verbal_prob_baseline}.
    The above process is carried out independently for each of the two additional conditions. The additional conditions with higher output probabilities for the gold outcome are the predicted answers. We use temperature = 0.7 to generate 3 responses and take the majority vote answer as the final answer. 
    \item \textbf{Logit-based Probability.} We directly use the output token probability for the decision. We use greedy decoding here to set the temperature = 0.
    \item \textbf{Chain-of-Thought Verbalization with Self-consistency (CoT).}~\citep{wei2022chain} We ask for a reasoning process to think step by step before verbalizing an estimated probability. The remaining process is the same as vanilla verbalization with SC. We show the Prompt in Appendix Fig.~\ref{tab:verbal_prob_baseline_1}.
    
    \item \textbf{Explicit Comparison (EC).} Instead of evaluating each additional condition separately, both conditions are presented simultaneously to the model. Given the scenario and the gold outcome, the model is asked to select which additional condition better supports the outcome without outputting any probabilities. Note that the task is less challenging for this particular baseline than the others due to the explicit formatting of inference in this baseline. As a result, performances under the EC setting are not directly comparable with our model and other baselines' performances since they only see one condition at a time and cannot perform argmax operations.  We show the Prompt in Appendix Fig.~\ref{tab:verbal_prob_baseline_explict}.

    \item \textbf{1/2 Assumption.} Instead of utilizing the learning algorithm mentioned in \S\ref{learning_algorithm} to estimate $\mathbb{P}(O_i|f)$. We adopt a neutral assumption. An outcome is determined based on the factors only if, except the values of the factors that are neutral, the values of all remaining factors uniformly support that particular outcome. Conversely, if the values of the factors indicate different outcomes and there is no prior information, we cannot conclusively determine the outcome.

\begin{small}
\begin{equation}
\mathbb{P}(O_i|f)  = \left\{ 
    \begin{array}{lc}
       1 & \mathbb{P}(O_i|f_j) \geq \frac{1}{2}, \forall j \ {\rm except} \ \mathbb{P}(O_i|f_j)=\frac{1}{2}, j=1,...,N\ \\
       \frac{1}{2} & {\rm otherwise} \\
    \end{array}
\right.
\end{equation}
\end{small}

The intuition behind this assumption is that as we focus solely on the present context, if the current information supports both outcomes and no additional information is available to finalize the decision, it becomes impossible to predict which outcome is more likely to occur.

\item \textbf{1/N Assumption.} Instead of utilizing the learning algorithm mentioned in \S\ref{learning_algorithm} to estimate $\mathbb{P}(O_i|f)$, we adopt another commonly used neutral assumption. The outcome $O_i$'s probability is determined by the aggregate count of factors whose values support outcome $O_i$.
\begin{equation}
\mathbb{P}(O_i|f)  = \frac{\sum_{j=1}^N \mathbbm{1}\{\mathbb{P}(O_i|f_j) > \frac{1}{2}\}}{N}
\end{equation}

\item \textbf{Fixed Initial Prob.} Instead of training with the algorithm to get an estimated prob for each $\mathbb{P}(O_i|f_j)$ in \S\ref{learning_algorithm}, we utilize the initial probability of Eq.\ref{eq:initial_prob} to directly estimate $\mathbb{P}(O_i|f)$ with Eq.\ref{eq:estimation}. 
\end{itemize}

\subsection{Analysis of Direct Decision Making}
\label{appendix:error}
We first conduct error analysis for Llama-2-70b-instruct on direct decision making where \framework{} gets the wrong outcomes while CoT makes the correct decisions. We detail the type of errors with its frequency percentage in Appendix Table~\ref{tab:error}. We observe that our proposed method outperforms the chain-of-thought baseline with Llama-3.1 rather than using Llama-2. This is partially because Llama-3.1 generates more diverse and accurate factors, and we present an example in Appendix Table~\ref{tab:error_example}.

\begin{table}[ht]
\centering
\small
\scalebox{1}{
{
\begin{tabular}{p{10cm}c} 
\toprule
\textbf{Error Type} &  \textbf{Frequency Percentage} \\
\cmidrule(lr){1-1}\cmidrule(lr){2-2}
The generated factors and their values are not comprehensive enough to cover all elements of the conditions. The selected values only imply partial information in the additional condition. & 56.7\% \\
\cmidrule(lr){1-2}
The additional condition is mapped to incorrect values of the factors that it does not necessarily imply. & 30\% \\
\cmidrule(lr){1-2}
The calculation of the final probability is inaccurate. & 13.3\% \\
\bottomrule
\end{tabular}
}}
\caption{Error analysis for Llama-2-70b-instruct for direct decision making.}
\label{tab:error}
\end{table}

\begin{table*}[htb]
\newcolumntype{?}{!{\vrule width 1pt}}
\newcolumntype{C}{>{\centering\arraybackslash}p{40em}}

\centering 
\renewcommand\arraystretch{1.0}
\small
\scalebox{0.95}{
\begin{tabular}{@{}l@{}}
\toprule
\textbf{Scenario}: the ease of carrying a cup is being compared between one person and six people. \\
\textbf{Outcome 1}: It would be easier for one person to carry the cup, than it would be for six people to carry the cup. \\
\textbf{Outcome 2}: It would be easier for six people to carry the cup, than it would be for one person to carry the cup.
 \\
\midrule
\textcolor{blue}{\textbf{Llama-2-70b-Instruct Generated Factors}} \\
\midrule
\textbf{Weight and Size of the Cup}:\\
The cup is light and small \\
The cup is heavy and large \\
The cup is of medium weight and size \\
\textbf{Material of the Cup}:\\

The cup is fragile \\

The cup is sturdy \\

The cup is made of moderately durable material \\

\textbf{Strength of the People}:\\
The person is able-bodied and strong enough to carry the cup \\
The combined strength of six people is greater than that of one person \\
\textbf{Experience of the Person or People Involved}: \\
The person has more experience in carrying  objects \\
The six people have more experience in lifting objects \\
\textbf{Formation of the People}: \\
The six people carry the cup in a specific formation designed to distribute the weight more evenly  \\

The six people do not carry the cup in a specific formation \\
\midrule
\textcolor{blue}{\textbf{Llama-3.1-70b-Instruct Generated Factors}} \\
\midrule
\textbf{Cup Size and Weight}: \\

The cup is small and light \\

The cup is large and heavy \\

The cup is small and heavy \\

The cup is large and light \\

\textbf{Cup Material}:\\

The cup is fragile \\

The cup is sturdy \\

The cup is constructed from a material of moderate durability\\

\textbf{Cup Design and Features}: \\

The cup has a comfortable grip and balanced ergonomics \\

The cup has an uncomfortable grip and unbalanced ergonomics \\

The cup has a neutral grip and average ergonomics \\
\textbf{Carrier's Physical Abilities}: \\

The one carrier has more strength and endurance than the six carries \\

The combined strength of the six carries is stronger than the one carrier \\

\textbf{Group Dynamics and Coordination}: \\

The group is well-coordinated and works together seamlessly \\

The group is disorganized and has difficulty working together \\

The group has average coordination and works together moderately well \\

\textbf{Physical Demands of the Task}: \\

The task requires exceptional strength and endurance \\

The task requires minimal strength and endurance \\

The task requires moderate strength and endurance \\

\textbf{The Size and Shape of the Carrying Area}: \\

The carrying area is narrow and confined \\

The carrying area is wide and open \\

\bottomrule
\end{tabular}
}
\caption{
	\label{tab:error_example} An example where Llama-3.1 generates better factors than Llama-2.
}
\end{table*}

\clearpage

We then give insights into LLM direct performance. We present an example from the \commonsense{} dataset in Appendix Table~\ref{tb:llm_example}. where the LLM accurately infers the correct outcome based predominantly on inductive biases, ignoring the other primary information provided in the condition, i.e., the peppers are used in large quantities, which will be captured in our framework.

\begin{table*}[htb]
\newcolumntype{?}{!{\vrule width 1pt}}
\newcolumntype{C}{>{\centering\arraybackslash}p{40em}}

\centering 
\renewcommand\arraystretch{1.0}
\small
\scalebox{0.95}{
\begin{tabular}{@{}l@{}}
\toprule
\textbf{Example} \\
\midrule
\textcolor{blue}{Scenario}: The spiciness of a hot sauce is being discussed in relation to the choice between using habanero \\ peppers or jalapeno peppers. \\
\textcolor{blue}{Condition}: The hot sauce recipe uses a significant amount of habanero peppers relative to the quantity \\of other ingredients.\\
\textcolor{blue}{Outcome 1}:  A hot sauce is going to be hotter if it uses habanero peppers rather than jalapeno.\\
\textcolor{blue}{Outcome 2}:  A hot sauce is going to be hotter if it uses jalapeno peppers rather than habanero.\\
\midrule
\textbf{Llama-2 Generated Deicision Making Process} \\
\midrule

\sethlcolor{pink}\hl{Habanero peppers are generally much hotter than jalapeno peppers. This is measured on the Scoville} \\
\sethlcolor{pink}\hl{scale, which is a measure of the pungency (spiciness or heat) of chili peppers and other spicy foods. So} \\
\sethlcolor{pink}\hl{the condition better supports Outcome 1: A hot sauce is going to be hotter if it uses habanero peppers}\\
\sethlcolor{pink}\hl{rather than jalapeno.}\\
\bottomrule
\end{tabular}
}
\caption{
	\label{tb:llm_example} An example where the LLM correctly makes the decision but with inadequate emphasis on the context.  
}
\end{table*}

\subsection{Ablation Study towards Abductive Factor Generation}
We compare the final decision-making accuracy based on two different factor generation methods and present the results in Appendix Table~\ref{tab:ablation}. We see that our proposed method performs much better than the direct generation method, indicating the effectiveness of utilizing the model's inductive bias inherited from training data. 

\begin{table}[ht]
\centering
\small
{
\begin{tabular}{lccccc}
\toprule
Dataset &  Method & $\framework{}_{llama2}$ & \%Unknown & $\framework{}_{llama3.1}$ & \%Unknown  \\
\cmidrule(lr){1-1}\cmidrule(lr){2-2}\cmidrule(lr){3-3}\cmidrule(lr){4-4}\cmidrule(lr){5-5}\cmidrule(lr){6-6}
\datasetname{}	& Direct	& 70.9 & 	74.1 &68.9& 46.3\\
	& Abduction	& 73.9 & 	54.0 & 74.3 & 34.0\\
 \cmidrule(lr){1-6}
\plasma{}& Direct  & 68.6 & 40.7 &69.0&24.1\\
  & Abduction & 74.0 & 31.4 &73.0&19.0\\
 \cmidrule(lr){1-6}
\cs{}{}{}{} & Direct & 84.2	& 54.9 &89.9&29.0\\
 & Abduction & 89.0 & 34.6 &92.3&27.1\\
\bottomrule
\end{tabular}
}
\caption{An ablation study towards different factor generation methods. Direct refers to the direct factor generation method. Abduction refers to our abductive factor generation.}
\label{tab:ablation}
\end{table}

\subsection{Ablation Study towards the effect of the two loss functions in the learning algorithm}
\label{appendix:loss}
We conduct an ablation study on the loss function in Eq.\ref{eq:loss} for decision making, as detailed in Appendix Table~\ref{tab:ablation_loss}, and demonstrate that both components of the loss function are crucial for final performance.

\begin{table}[ht]
\centering
\small
{
\begin{tabular}{lccc}
\toprule
Dataset &  w/o MR Loss	 & w/o MSE Loss & 	BIRD \\
\cmidrule(lr){1-1}\cmidrule(lr){2-2}\cmidrule(lr){3-3}\cmidrule(lr){4-4}
\datasetname{}	& 65.5	&65.2	&74.3\\

\plasma{}& 	60.0&	65.7	&73.0 \\

\cs{} & 77.2	&86.8	&92.3\\
\bottomrule
\end{tabular}
}
\caption{An ablation study towards the effect of the two loss functions in Eq.\ref{eq:loss} for decision making with Llama-3.1-70b-Instruct.}
\label{tab:ablation_loss}
\end{table}

\subsection{Experiment Settings for Using Estimated Probabilities as additional training data}
\label{appendix:probabilitytrain}
The T5-large model and PatternTime~\citep{zhou-etal-2021-temporal} are specialized temporal reasoning models, as outlined in~\cite{feng-etal-2023-generic}. Both models were specifically fine-tuned on temporal reasoning datasets using the original temporal training data from~\cite{feng-etal-2023-generic}. In contrast, the additional training data from \commonsense{} in \S\ref{extrinsic} focuses on commonsense reasoning.

\subsection{Yes-or-no followup Question Generation}
\label{appendix:yesorno}
We experiment to see if \framework{} can further facilitate decision making in a controllable way, where we use \framework{} to generate a yes-or-no follow-up question about an unobserved factor in the context. Specifically, given a scenario and an additional condition, we first identify which factors and their corresponding values are implied by the condition, we then select a factor that is not mentioned and is most important for the decision in the remaining unmentioned factors, as the target question area. We adopt the factor value that better aligns with the outcome direction suggested by the current implied factor values and has the largest condition probability. We then rewrite the statement of the factor value into its corresponding question. Simultaneously, we ask Llama2-70b-instruct to directly generate a yes-or-no follow-up question. The prompt is in Appendix Fig.~\ref{fig:question}. 

We use MTurk and ask annotators to provide pairwise preference between the \framework{} generated question and LLM directly generated question. We adopt the same Mturk setting as mentioned in \S\ref{appendix:humaneval}. we carry out the human annotation process three times, using the majority vote to find the preferred generated question. We show the experiment interfaces and an example in Appendix Fig.~\ref{fig:human_annotation_1}. Appendix Table~\ref{tab:question} shows that the follow-up questions generated by \framework{} are more often preferred.

 \begin{figure*}[ht]
\centering
\scalebox{1}{
\includegraphics[width=0.98\textwidth]{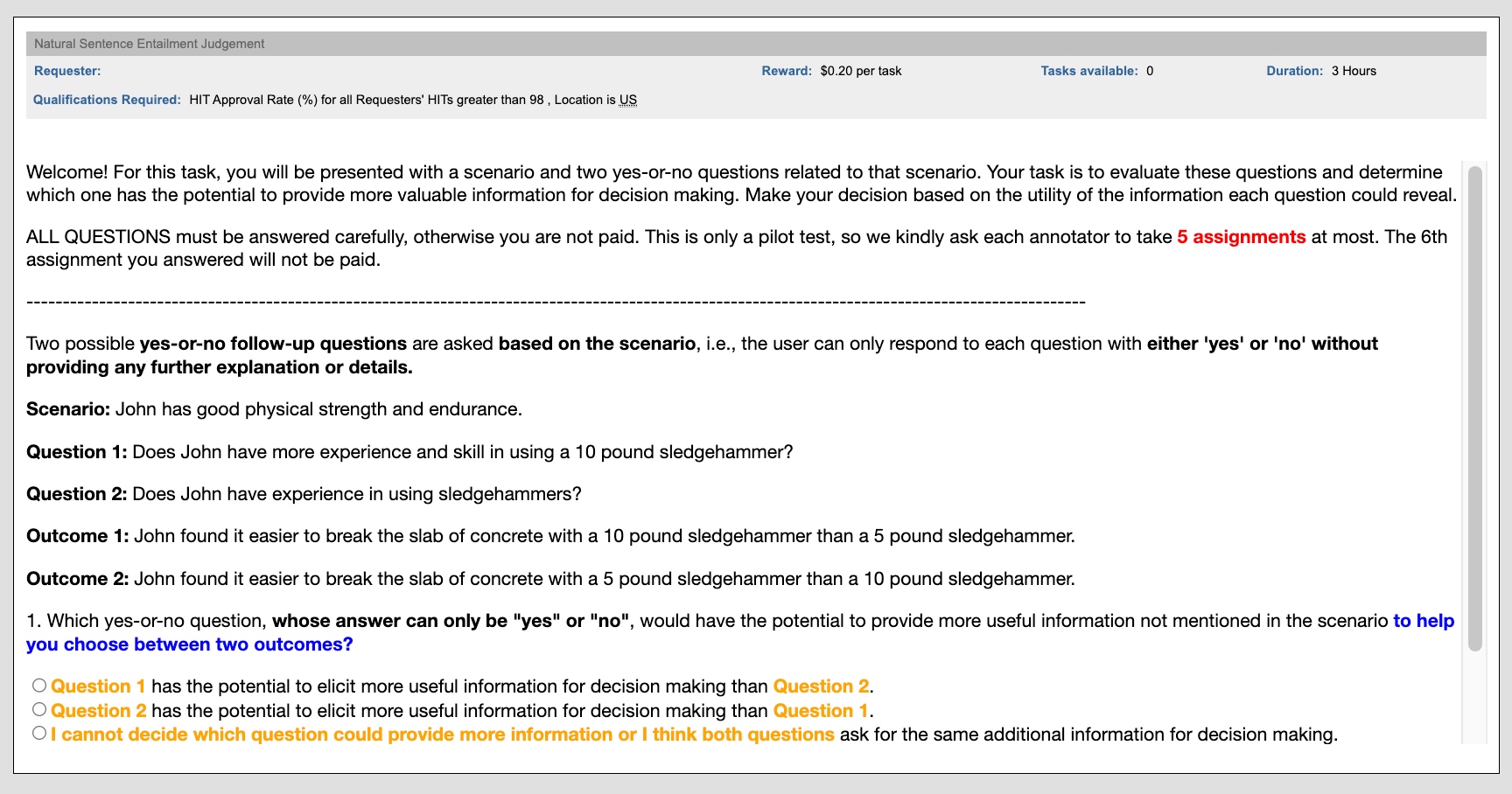}}
	\caption{\label{fig:human_annotation_1}The interface for human evaluation on preference-based pairwise evaluation of the generated questions. Question 1 is generated by \framework{} and question 2 is generated by LLM directly. The three annotators all prefer question 1. }
\end{figure*}

\begin{table}[ht]
\centering
{
\begin{tabular}{lccc}
\toprule
Dataset & \framework{} Win Direct & Tie & \framework{} Lose Direct \\
\cmidrule(lr){1-1}\cmidrule(lr){2-2}\cmidrule(lr){3-3}\cmidrule(lr){4-4}
\commonsense{} &132	&36	&82
 \\
\bottomrule
\end{tabular}
}
\caption{Preference judgments for yes-or-no follow-up questions generated by \framework{} versus questions directly generated by LLMs. We randomly sample 250 instances from \commonsense{}.} 
\label{tab:question}
\end{table}

\subsection{Prompt}

\newtcolorbox{CoTAgBox_2}[1][]{%
  floatplacement=htbp,   
  float,
  colframe=blue!60!black,  %
  colback=blue!5!white,
  title=\textsc{Example \textbf{Additional Sentence Sampling} Prompt},
  #1
}

\begin{CoTAgBox_2}
\vspace{1mm}
{

\tt \small  
\textsc{\textbf{\underline{System}}}

You are given a scenario and an accompanying hypothesis. Generate 5 sentences covering different conditions that would add objective information relevant to the hypothesis such that the hypothesis is more likely to hold true. The information should not definitively imply the hypothesis. You must follow the below structure to just generate sentences with no explanations.

\tt \small  
\textsc{\textbf{\underline{User}}}

Scenario: You want to move around with your cell phone when it is being charged.

Hypothesis:  You can move around more freely with your cell phone if it is being charged with a one-foot cord rather than a six-foot cord.

\tt \small  
\textsc{\textbf{\underline{Assistant}}}

\# The cell phone is being charged with a portable power bank located in your pocket, allowing you to move around without being tethered to a fixed outlet.

\# The user is working in a compact space where longer cords could easily snag on furniture or equipment, thus a one-foot cord could minimize this risk.

\# The phone is needed for tasks that require frequent handling and close proximity to the user, making a shorter cord more practical to avoid excessive dangling.

\# The charging setup includes a small desktop charger that keeps the phone elevated and stable, limiting the practicality of a longer cord.

\# The user is in a busy environment like a kitchen or workshop, where shorter cords can reduce the hazard of tripping or catching on moving objects.

}
\captionof{figure}{Example \textbf{Additional Sentence Sampling} Prompt}\label{fig:factor_added_info}
\vspace{0.5mm}
\end{CoTAgBox_2}

\newtcolorbox{CoTAgBox_3}[1][]{%
  floatplacement=htbp,   
  float,
  colframe=blue!60!black,  %
  colback=blue!5!white,
  title=\textsc{Example \textbf{Factor Summarization from Additional Sentences} Prompt},
  #1
}

\begin{CoTAgBox_3}
\vspace{1mm}
{

\tt \small  
\textsc{\textbf{\underline{System}}}

From the given sentences for each outcome, identify and list distinct and concrete factors, ensuring each is broad yet specific and focuses on a unique aspect. 

Your response should strictly adhere to the JSON format provided, without additional explanations.

For example:
\{"distinct factor" <ENSURE each factor focuses on a unique aspect>: "factor values" <Each factor MUST cover at least one condition to support the Statement and one condition to support the Opposite statement.>\}

1. Ensure that each factor's value MUST directly reference specific elements mentioned in the statements, avoiding vague terms like 'the object'. 

2. Ensure the factor values are not too concrete.

3. Do not only mention the common situations.

\tt \small  
\textsc{\textbf{\underline{User}}}

Scenario: You are charging your cell phone and wish to move around with your cell phone. 

Outcome 1: You can move around more freely with your cell phone if it is being charged with a six feet cord rather than a one feet cord.

Sentences:

\#1 A longer cord provides more flexibility and allows for a greater range of movement while using the cell phone. This is because the additional length of the six-foot cord gives the user a larger radius of movement, enabling them to comfortably use their phone while it is charging without feeling restricted or confined to a specific location.

Outcome 2: You can move around more freely with your cell phone if it is being charged with a one-foot cord rather than a six-foot cord.

Sentences:

\#2 If the cell phone is plugged into a portable power bank or a USB port on a computer, a one-foot cord provides greater mobility because it is shorter and less likely to get tangled or caught on objects while moving.

\#3 If the cell phone is constantly being used while charging and the user prefers to keep the phone close to the charger at all times, a one-foot cord allows for easier mobility and reduces the risk of tripping over a longer cord."

\tt \small  
\textsc{\textbf{\underline{Assistant}}}

\{
    "The cell phone's charging method": 
    
    ["The charger is portable", "The charger is unmovable"],
    
    "The user's movement range": 
    
    ["The user stays very close to the charger", "The user has a large radius of movement"],
    
    "The location of the phone charger": 
    
    ["The user leaves the charger somewhere", "The user carries the charger"]\}

}
\captionof{figure}{Example \textbf{Factor Summarization from Additional Sentences} Prompt}\label{fig:factor_structure}
\vspace{0.5mm}
\end{CoTAgBox_3}

\newtcolorbox{CoTAgBox_4}[1][]{%
  floatplacement=htbp,   
  float,
  colframe=blue!60!black,  %
  colback=blue!5!white,
  title=\textsc{Example Prompt for \textbf{Deciding the Value Better Supports Which Outcome} },
  #1
}

\begin{CoTAgBox_4}
\vspace{1mm}
{

\tt \small  
\textsc{\textbf{\underline{System}}}

Decide whether each of the given values supports Statement 1 or Statement 2. The statement and opposite statement are provided, along with a list of values. Determine which statement each value supports. You must follow the below structure to just generate json with an explanation.

For example:

Explanation: <explanations for how each value supports which statement>

List: \{
"value 1":<output only Statement 1, Statement 2, Both, or Neither>,
"value 2":<output only Statement 1, Statement 2, Both, or Neither>\}

\tt \small  
\textsc{\textbf{\underline{User}}}

Here is a scenario: ... I'd like you to determine which statement each value supports.

Values: ...

* Statement 1: ...

* Statement 2: ...
                        
Please analyze each value and determine which statement it supports.

\tt \small  
\textsc{\textbf{\underline{Assistant}}}

Explanation: [step-by-step analysis here]

List: 
\{
"value":"Statement 1"
\}

}
\captionof{figure}{Example Prompt for \textbf{Deciding the Value Better Supports Which Outcome} }\label{fig:factor_outcome}
\vspace{0.5mm}
\end{CoTAgBox_4}

\newtcolorbox{CoTAgBox_5}[1][]{%
  floatplacement=htbp,   
  float,
  colframe=blue!60!black,  %
  colback=blue!5!white,
  title=\textsc{Example Prompt for \textbf{Deciding If the Scenario with the Condition Implies the Value of the Factor} },
  #1
}

\begin{CoTAgBox_5}
\vspace{1mm}
{

\tt \small  
\textsc{\textbf{\underline{System}}}

Given a scenario, read a sentence and determine whether a value from a given list is implied by the sentence. You must be strict on considering a value as implied and return one value or "None" with an explanation.

Here are rules that you must follow:

1. A value is implied only if the sentence implies it in most situations, meaning the sentence provides clear, direct information relevant to the value.

2. Do not select a value based on indirect associations, context, or related hints unless they strongly imply the value in most situations.

3. Do not assume causality or infer a connection if it is not clearly stated in the sentence.

4. If the sentence implies more than one value, choose the most implied one. If two values are both implied, choose the one that is positive (non-negative).

5. If the sentence does not imply any value directly in most situations, say "None".

\tt \small  
\textsc{\textbf{\underline{User}}}

Here is a scenario: ...

Sentence: ...

List: "factor": [
                "value 1", ...
                ]

\tt \small  
\textsc{\textbf{\underline{Assistant}}}

\{"explanation": ...,

"answer": a value/None\}
                
}
\captionof{figure}{Example Prompt for \textbf{Deciding If the Scenario with the Condition Implies the Value of the Factor}}\label{fig:entailment}
\vspace{0.5mm}
\end{CoTAgBox_5}

\newtcolorbox{CoTAgBox_6}[1][]{%
  floatplacement=htbp,   
  float,
  colframe=blue!60!black,  %
  colback=blue!5!white,
  title=\textsc{Example Prompt for \textbf{Reevaluating Implied Value} },
  #1
}

\begin{CoTAgBox_6}
\vspace{1mm}
{

\tt \small  
\textsc{\textbf{\underline{System}}}

Based on a scenario, delete value in the list if the key does not imply value. You should output an explanation and then a json. You must generate double quotes.

Here are rules that you must follow:

- Ensure that retained values align with the key's implications, do not assume causality or infer a connection if the value is not clearly stated in the key. Avoid over-generalizations and ensure logical consistency in your decisions.

- Be lenient with deletions. Only remove a value if the key clearly does not imply the value.

- Provide a clear explanation for each deletion or retention.

\tt \small  
\textsc{\textbf{\underline{User}}}

Scenario: The government is planning the locations for building charging stations.

\{"The location is on a busy highway with no existing charging stations.": 

["No nearby charging stations",

"Location is on a major travel route, serving long-distance EV travelers",

"Nearby amenities like restaurants, shops, and rest areas for users while charging"]\}

\tt \small  
\textsc{\textbf{\underline{Assistant}}}

Explanation: "No nearby charging stations" was kept because the key explicitly mentions the absence of existing charging stations in the area. Similarly, "Location is on a major travel route, serving long-distance EV travelers" was retained because being on a busy highway strongly implies that it serves as a key travel route for such travelers. However, "Nearby amenities like restaurants, shops, and rest areas for users while charging" was removed since the key does not directly mention or suggest the presence of these amenities.

List: \{"The location is on a busy highway with no existing charging stations.": 

["No nearby charging stations",

"Location is on a major travel route, serving long-distance EV travelers"]\}

}
\captionof{figure}{Example Prompt for \textbf{Reevaluating Implied Value}. Explanation is optional. }\label{fig:entailment_reeval}
\vspace{0.5mm}
\end{CoTAgBox_6}

\newtcolorbox{CoTAgBox_7}[1][]{%
  floatplacement=htbp,   
  float,
  colframe=blue!60!black,  %
  colback=blue!5!white,
  title=\textsc{Example Prompt for \textbf{Outputting a Verbalized Probability Score} },
  #1
}

\begin{CoTAgBox_7}
\vspace{1mm}
{

\tt \small  
\textsc{\textbf{\underline{System}}}

As an AI assistant, your role is to respond accurately to user queries. While answering think step-by-step and justify your answer.
Analyze the given scenario and condition to determine the likelihood of the outcomes. Use only the information provided, without relying on prior knowledge. Assess the probability using the specified terms: 'very likely', 'likely', 'somewhat likely', 'neutral', 'somewhat unlikely', 'unlikely', 'very unlikely'. Ensure that your assessments are complementary: if one outcome is deemed 'likely', the other must be 'unlikely', and so on.  You should first give your reasons and then format your final answer.

\tt \small  
\textsc{\textbf{\underline{User}}}

Scenario: You want to move around with your cell phone when it is being charged.

Condition: The charger is portable. The user stays very close to the charger. The user carries the charger.

Outcome 1: You can move around more freely with your cell phone with a one-foot cord rather than a six-foot cord.

Outcome 2: You can move around more freely with your cell phone with a six-foot cord rather than a one-foot cord.

\tt \small  
\textsc{\textbf{\underline{Assistant}}}

Given that the user is carrying a portable charger, a shorter cord like one foot would indeed be far more manageable, making it easier for the user to move around freely.

Given the same conditions, a longer cord like six feet might become an impediment, making it more challenging for the user who is carrying the charger to move around freely due to the possibility of tangling or managing the extra length.

Therefore, 
Outcome 1: Likely

Outcome 2: Unlikely

}
\captionof{figure}{Example Prompt for \textbf{Outputting a Verbalized Probability Score} }\label{fig:llm_training_signal}
\vspace{0.5mm}
\end{CoTAgBox_7}

\newtcolorbox{CoTAgBox_8}[1][]{%
  floatplacement=htbp,   
  float,
  colframe=blue!60!black,  %
  colback=blue!5!white,
  title=\textsc{Example Prompt for \textbf{Direct Probability Verbalization Baseline} },
  #1
}

\begin{CoTAgBox_8}
\vspace{1mm}
{

\tt \small  
\textsc{\textbf{\underline{System}}}

You are an AI assistant that helps people make decisions. The user will give you a question. Choose which outcome the scenario and its corresponding additional condition better support. Provide your confidence level as a percentage to indicate how certain you are about your answer.

\tt \small  
\textsc{\textbf{\underline{User}}}

Scenario: … 

Additional Condition: … 

Outcome 1: … 

Outcome 2: … 

\tt \small  
\textsc{\textbf{\underline{Assistant}}}

Answer and Confidence (0-100): Outcome i, X%

}
\captionof{figure}{Example Prompt for \textbf{Direct Probability Verbalization Baseline} }\label{tab:verbal_prob_baseline}
\vspace{0.5mm}
\end{CoTAgBox_8}

\newtcolorbox{CoTAgBox_9}[1][]{%
  floatplacement=htbp,   
  float,
  colframe=blue!60!black,  %
  colback=blue!5!white,
  title=\textsc{Example Prompt for \textbf{Direct Probability Verbalization with COT Baseline} },
  #1
}

\begin{CoTAgBox_9}
\vspace{1mm}
{

\tt \small  
\textsc{\textbf{\underline{System}}}

You are an AI assistant that helps people make decisions. The user will give you a question. Your task is to answer as faithfully as you can. While answering, think step-by-step and justify your answer. You are asked to provide a probability for each outcome being correct (0\% to 100\%) based on the scenario and its corresponding additional condition. Ensure that P(outcome 1) + P(outcome 2) = 1 .

\tt \small  
\textsc{\textbf{\underline{User}}}

Scenario: … 

Additional Condition: … 

Outcome 1: … 

Outcome 2: … 

\tt \small  
\textsc{\textbf{\underline{Assistant}}}

Explanation: [step-by-step analysis here]

Confidence: P(outcome 1) = X\%, P(outcome 2) = Y\%.

}
\captionof{figure}{Example Prompt for \textbf{Direct Probability Verbalization with CoT Baseline} }\label{tab:verbal_prob_baseline_1}
\vspace{0.5mm}
\end{CoTAgBox_9}

\newtcolorbox{CoTAgBox_10}[1][]{%
  floatplacement=htbp,   
  float,
  colframe=blue!60!black,  %
  colback=blue!5!white,
  title=\textsc{Example Prompt for \textbf{Explicit Comparison Baseline} },
  #1
}

\begin{CoTAgBox_10}
\vspace{1mm}
{

\tt \small  
\textsc{\textbf{\underline{System}}}

You are an AI assistant that helps people make decisions. The user will give you a question. Your task is to answer as faithfully as you can. While answering, think step-by-step and justify your answer. You are asked to decide which condition makes the outcome more likely to be true given the scenario.

\tt \small  
\textsc{\textbf{\underline{User}}}

Scenario: … 

Outcome: … 

Condition 1: … 

Condition 2: … 

\tt \small  
\textsc{\textbf{\underline{Assistant}}}

Explanation: [step-by-step analysis here]

Answer: Condition i/Both.

}
\captionof{figure}{Example Prompt for \textbf{Explicit Comparison Baseline} }\label{tab:verbal_prob_baseline_explict}
\vspace{0.5mm}
\end{CoTAgBox_10}

\newtcolorbox{CoTAgBox}[1][]{%
  floatplacement=htbp,   
  float,
  colframe=blue!60!black,  %
  colback=blue!5!white,
  title=\textsc{Example \textbf{COT for Decision Making} Prompt},
  #1
}

\begin{CoTAgBox}
\vspace{1mm}
{

\tt \small  
\textsc{\textbf{\underline{System}}}

You are a helpful assistant. While answering think step-by-step and justify your answer. Given a scenario and an additional condition, decide which of the two outcomes is supported. 

\tt \small  
\textsc{\textbf{\underline{User}}}

Scenario: The government is planning the locations for building charging stations.

Condition: The location is on a busy highway with no existing charging stations.

Outcome 1: The government should build a charging station here.

Outcome 2: The government should not build a charging station here.

\tt \small  
\textsc{\textbf{\underline{Assistant}}}

The scenario presents a busy highway with no existing charging stations, and the government is planning where to build new ones. Given the high traffic volume and the absence of charging infrastructure, building a station here would fill a critical gap, support electric vehicle drivers, and encourage the transition to sustainable transportation.

Based on the available information, Outcome 1: The government should build a charging station here is more supported.

}
\captionof{figure}{Example \textbf{COT for Decision Making} Prompt}\label{fig:prompt-cot}
\vspace{0.5mm}
\end{CoTAgBox}

\newtcolorbox{CoTAgBox_1}[1][]{%
  floatplacement=htbp,   
  float,
  colframe=blue!60!black,  %
  colback=blue!5!white,
  title=\textsc{Example \textbf{Direct Yes-or-no Follow-up Question Generation} Prompt},
  #1
}

\begin{CoTAgBox_1}
\vspace{1mm}
{

\tt \small  
\textsc{\textbf{\underline{System}}}

You are a helpful assistant. While answering think step-by-step and justify your answer. Ask a follow-up yes or no question so that you can better decide which outcome is better supported by the scenario and the condition. You can only ask for a specific situation.

\tt \small  
\textsc{\textbf{\underline{User}}}

Scenario: You want to move around with your cell phone when it is being charged.

Condition: The charger is portable and the user carries the charger.

Outcome 1: You can move around more freely with your cell phone with a one-foot cord rather than a six feet cord.

Outcome 2: You can move around more freely with your cell phone with a six-foot cord rather than a one feet cord.

\tt \small  
\textsc{\textbf{\underline{Assistant}}}

In this scenario, the primary factor to consider is cord length, in most situations, a greater cord length allows more movement freedom, as you can be further from the outlet and still keep your device charged. However, considering that the user carries the charger, it might bring about the confusion between the two outcomes. 

Thus, to further ascertain which outcome is supported better, I'd ask the following question: 

Does the portable charger need to be plugged into an outlet to charge the phone?

}
\captionof{figure}{Example \textbf{Direct Yes-or-no Follow-up Question Generation} Prompt}\label{fig:question}
\vspace{0.5mm}
\end{CoTAgBox_1}

\end{document}